\documentclass[runningheads]{llncs}

\usepackage{eccv}

\usepackage{eccvabbrv}

\usepackage{graphicx}
\usepackage{booktabs}

\usepackage[dvipsnames]{xcolor}

\usepackage{amsmath}
\usepackage{amssymb}
\usepackage[linesnumbered, ruled, vlined]{algorithm2e}
\SetKwInput{KwData}{In}
\SetKwInput{KwResult}{Out}
\SetAlCapNameFnt{\footnotesize}
\SetAlCapFnt{\footnotesize}

\usepackage{dsfont}
\usepackage{bm}
\usepackage{float}
\usepackage{subcaption}
\usepackage{capt-of}

\usepackage{multirow}
\usepackage{tikz}
\usepackage{pgfplots}
\pgfplotsset{width=10cm,compat=1.9}
\usepgfplotslibrary{groupplots}
\usepgfplotslibrary{fillbetween}

\usepackage{dblfloatfix}
\usepackage{url}

\usetikzlibrary{positioning}
\usetikzlibrary{shapes.geometric, positioning, calc, fit}
\usetikzlibrary{decorations.text}

\usepackage{arydshln}

\makeatletter
\def\adl@drawiv#1#2#3{%
        \hskip.5\tabcolsep
        \xleaders#3{#2.5\@tempdimb #1{1}#2.5\@tempdimb}%
                #2\z@ plus1fil minus1fil\relax
        \hskip.5\tabcolsep}
\newcommand{\cdashlinelr}[1]{%
  \noalign{\vskip\aboverulesep
           \global\let\@dashdrawstore\adl@draw
           \global\let\adl@draw\adl@drawiv}
  \cdashline{#1}
  \noalign{\global\let\adl@draw\@dashdrawstore
           \vskip\belowrulesep}}
\makeatother

\newcommand{\spm}[1]{{\tiny$\pm$#1}}
\DeclareMathOperator*{\argmax}{argmax}

\DeclareMathOperator*{\R}{\mathbb{R}}

\newcommand\tightdots{\hbox to 0.7em{.\hss.\hss.}}

\AtBeginDocument{%
\crefname{equation}{}{}
\Crefname{equation}{}{}
\crefrangeformat{equation}{(#3#1#4)--(#5#2#6)}
}

\usepackage[accsupp]{axessibility}  %

\usepackage{hyperref}

\usepackage{orcidlink}

\begin{document}

\title{ProSub: Probabilistic Open-Set Semi-Supervised Learning with Subspace-Based Out-of-Distribution Detection} 

\titlerunning{ProSub}

\author{Erik Wallin\inst{1,2} \and
Lennart Svensson\inst{2} \and
Fredrik Kahl\inst{2} \and
Lars Hammarstrand\inst{2}}

\authorrunning{E.~Wallin et al.}

\institute{\textsuperscript{1}Saab AB, \textsuperscript{2}Chalmers University of Technology
\email{\{walline,lennart.svensson,fredrik.kahl,lars.hammarstrand\}@chalmers.se}}

\maketitle

\begin{abstract}

In open-set semi-supervised learning (OSSL), we consider unlabeled datasets that may contain unknown classes. Existing OSSL methods often use the softmax confidence for classifying data as in-distribution (ID) or out-of-distribution (OOD). Additionally, many works for OSSL rely on ad-hoc thresholds for ID/OOD classification, without considering the statistics of the problem. We propose a new score for ID/OOD classification based on angles in feature space between data and an ID subspace. Moreover, we propose an approach to estimate the conditional distributions of scores given ID or OOD data, enabling probabilistic predictions of data being ID or OOD. These components are put together in a framework for OSSL, termed \emph{ProSub}, that is experimentally shown to reach SOTA performance on several benchmark problems. Our code is available at \url{https://github.com/walline/prosub}. 

\end{abstract}
\section{Introduction}
\label{sec:intro}

Open-set semi-supervised (OSSL) learning is the realistic setting of semi-super\-vised learning in which we \emph{do not assume} that the unlabeled data only contain the classes of interest (the classes in the labeled set) \cite{guo2020safe, chen2020semi, yu2020multi, saito2021openmatch}. This setting is of practical importance since one of the advantages of unlabeled data lies in its freedom from human vetting, thus making it hard to ensure that the data only contain known classes. Moreover, if data with unknown classes appear during training, similar data may likely appear at test time, making it essential to identify these data in deployment.

Many existing methods %
enable learning from unlabeled data through some form of pseudo-labeling: assigning artificial training labels to unlabeled samples through model predictions. A key challenge of this approach is to assign sufficiently many correct pseudo-labels to unlabeled data to effectively learn to classify the ID classes, without incorrectly assigning pseudo-labels to OOD samples, which can harm the model performance for ID/OOD detection. To this end, an accurate method to separate ID and OOD in training is crucial for OSSL.

A common approach for separating ID and OOD %
is to employ the maximum softmax probability \cite{chen2020semi, huang2022they, han2023pseudo}, the idea being that unlabeled data that are ID tend to yield larger confidences than OOD data. While the maximum softmax probability can act as a strong baseline, many works outside the domain of OSSL have proposed stronger scores for ID/OOD classification \cite{lee2018simple, liu2023gen, wang2022vim}. This suggests the existence of better-performing alternatives %
in the context of OSSL. Additionally, many methods for OSSL rely on ad-hoc thresholds for ID/OOD classification \cite{saito2021openmatch, park2022opencos, han2023pseudo, he2022safe} that do not adapt to the difficulty of the problem or the learning status of the model. Combined, these drawbacks may lead to inaccuracies and over- or under-confidence in classifying data as ID or OOD. 

\begin{figure}[!t]
    \centering
    \input{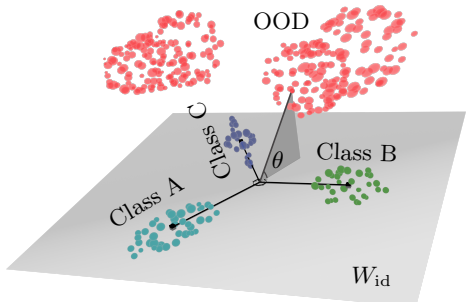}
    \caption{Our ID subspace, $W_\text{id}$, spanned by the class centers. The angle, $\theta$, to this space is generally larger for OOD data than for ID data and is used as a score in ProSub.}
    \label{fig:firstpagefig}
\end{figure}

To address these limitations of existing works, we propose two new components for OSSL. Firstly, we suggest a novel score for classifying data as ID or OOD. Work on the phenomenon of neural collapse has found that features of labeled data converge toward class centers in the output space of the penultimate network layer \cite{papyan2020prevalence}. Based on this observation, we propose the notion of an ID subspace as the space spanned by the class means in this feature space. Coupled with training using cosine-based self-supervision, we find that evaluating the angles between features of data and this subspace presents a strong score for ID/OOD classification in OSSL (see \cref{fig:firstpagefig}). Additionally, the distributions of this score given ID or OOD data have the advantage of being well-modeled by two Beta distributions.

Secondly, to avoid relying on manually set thresholds for ID/OOD classification, we estimate this pair of Beta distributions. With accurate density models, we can obtain probabilistic predictions for samples being ID or OOD. For this estimation, we propose an approach inspired by the expectation-maximization (EM) algorithm \cite{dempster1977maximum}, in which samples being ID or OOD is an unobserved (hidden) variable for unlabeled data.
Additionally, to fully utilize the probabilistic predictions, we use a procedure where hard binary pseudo-labels for ID or OOD are sampled based on the predicted probabilities.

Finally, we combine these components to form a framework for OSSL, \emph{ProSub}, and demonstrate through experimental evaluation that this method achieves state-of-the-art results on closed-set accuracy and AUROC for classifying data as ID or OOD on many benchmark problems.

\newpage
The main contributions of this work are:
\begin{itemize}
    \item \emph{ProSub}, a framework for OSSL achieving state-of-the-art results on several benchmarks.
    \item An ID/OOD score %
    based on the angle in feature space to an ID subspace.
    \item An adaptive approach to enable probabilistic ID/OOD predictions, %
    achieved by estimating the conditional distributions of scores given ID or OOD data through an iterative algorithm.
\end{itemize}

\section{Related work}
\label{sec:related-work}

\textbf{Semi-supervised Learning (SSL):} In SSL, we use training data where only part of the data have labels \cite{lee2013pseudo, rasmus2015semi, laine2016temporal, tarvainen2017mean}. A large part of the works in SSL consider the closed-set setting, where we assume the unlabeled data contain the same classes as the labeled data. Currently, most methods use different forms of pseudo-labeling and consistency regularization  \cite{xu2021dash, zhang2021flexmatch, yang2022class, nassar2023protocon, wang2022freematch, zheng2022simmatch, wang2022np, xie2019unsupervised}, using augmentation strategies involving weak and strong augmentations introduced in \cite{berthelot2019remixmatch, sohn2020fixmatch}. For ProSub, we adopt this widely used augmentation strategy for unlabeled data. We also include a pseudo-labeling procedure similar to FixMatch \cite{sohn2020fixmatch} and the self-supervised component proposed by DoubleMatch \cite{wallin2022doublematch}.

\textbf{Open-Set Semi-supervised Learning:} OSSL relaxes the closed-set assumption of SSL and considers unlabeled data that can contain unknown classes, not present in the labeled data \cite{chen2020semi, guo2020safe, he2022safe, yu2020multi, saito2021openmatch, huang2021trash, huang2022they, huang2023fix, he2022not, wang2023out, park2022opencos, mo2023ropaws, han2023pseudo, zhao2022out, wallin2023improving, li2023iomatch, fan2023ssb, ma2023rethinking}. Some works focus only on obtaining a high accuracy on the closed set (closed-set accuracy) \cite{han2023pseudo, huang2023fix, mo2023ropaws, chen2020semi}, whereas other works focus on both high closed-set accuracy and accurate ID/OOD classification \cite{saito2021openmatch, huang2021trash, yu2020multi, wallin2023improving}. Many early works for OSSL adopted an approach where OOD data are rejected from unlabeled data and remaining data are included in a (closed-set) SSL loss \cite{yu2020multi, chen2020semi, guo2020safe}. More recent works have found it beneficial to enable learning signals from all unlabeled data, whether ID or OOD, from, \eg, self-supervision \cite{huang2021trash, wallin2023improving} or pseudo-labeling where also OOD data are included \cite{li2023iomatch, fan2023ssb}.

While we are (to our knowledge) first to introduce an adaptive and probabilistic approach for classifying unlabeled data as ID or OOD in OSSL, existing methods have explored adaptive thresholds. For example, MTCF \cite{yu2020multi}, T2T \cite{huang2021trash}, and OSP \cite{wang2023out} resort to Otsu thresholding \cite{otsu1979threshold} to determine a threshold based on the scores of unlabeled data. The Otsu algorithm is originally a method for classifying the pixels of an image into background and foreground. While this method avoids the need for a manually determined threshold, the resulting binary classifier does not capture the uncertainty of the problem.

UASD \cite{chen2020semi} proposes to adaptively change the threshold based on the average confidence on a labeled validation set. While this method successfully adapts to the current confidence of the model, it does not consider statistics of OOD data, and the resulting classifier is binary. Similarly, SeFOSS \cite{wallin2023improving} proposes a method to compute energy score thresholds based on the labeled training data statistics. Our proposed model considers the statistics of both ID and OOD data and yields a probabilistic prediction of each sample being ID or OOD.

A setting similar to OSSL is open-world SSL, which expands the classification problem to include unknown classes in unlabeled data \cite{cao2022open, rizve2022towards, rizve2022openldn, liu2023open}. Another related field is long-tailed SSL, which studies SSL under class imbalances \cite{wei2021crest, kim2020distribution, wei2023towards}, but typically does not assume the presence of unknown classes.

\textbf{Open-Set Recognition:} Predicting if data belong to a pre-defined set of classes is often referred to as open-set recognition (OSR) or OOD detection. This problem occurs naturally as part of OSSL but is also widely studied in a broader context \cite{scheirer2012toward, bendale2016towards, hendrycks2016baseline, hendrycks2018deep, lakshminarayanan2017simple, liang2017enhancing, wang2022vim, liu2023gen}. Recently, methods for OOD detection based on measuring distances to ID training data in some feature space \cite{lee2018simple, ming2023cider, sehwag2021ssd, sun2022out} have gained a lot of traction as an improvement to confidence-based methods \cite{hendrycks2016baseline}.

In ProSub, we build upon the idea of distance-based OOD detection and use the notion of an ID subspace, $W_\text{id}$, in feature space. Similar ideas are explored in Vim \cite{wang2022vim} and concurrently to us in Neco \cite{ammar2024neco}, both utilizing ID subspaces for OOD detection. Vim assesses ID/OOD-ness by computing the residual of a test vector's projection onto such a space, whereas Neco, similarly to us, evaluates the angle to this space. However, Vim and Neco use PCA of the features for the full training set to compute the ID space which would be too expensive in an OSSL setting where we need accurate OOD predictions during the entire training process. In contrast, we use a cheap method for computing $W_\text{id}$ continuously during training based on the class means of labeled data, better suited for OSSL.

Furthermore, Vim and Neco use additional operations to scale their scores with the predicted logits. For ProSub, we empirically find that in conjunction with the self-supervision from \cite{wallin2022doublematch}, using the cosine of the angle to $W_\text{id}$ directly offers the dual benefits of strong OSR and a good fit with the Beta distribution. %

\section{Model}
\label{sec:method}

\begin{figure*}[!t]
    \centering
    \includegraphics[width=0.53\linewidth]{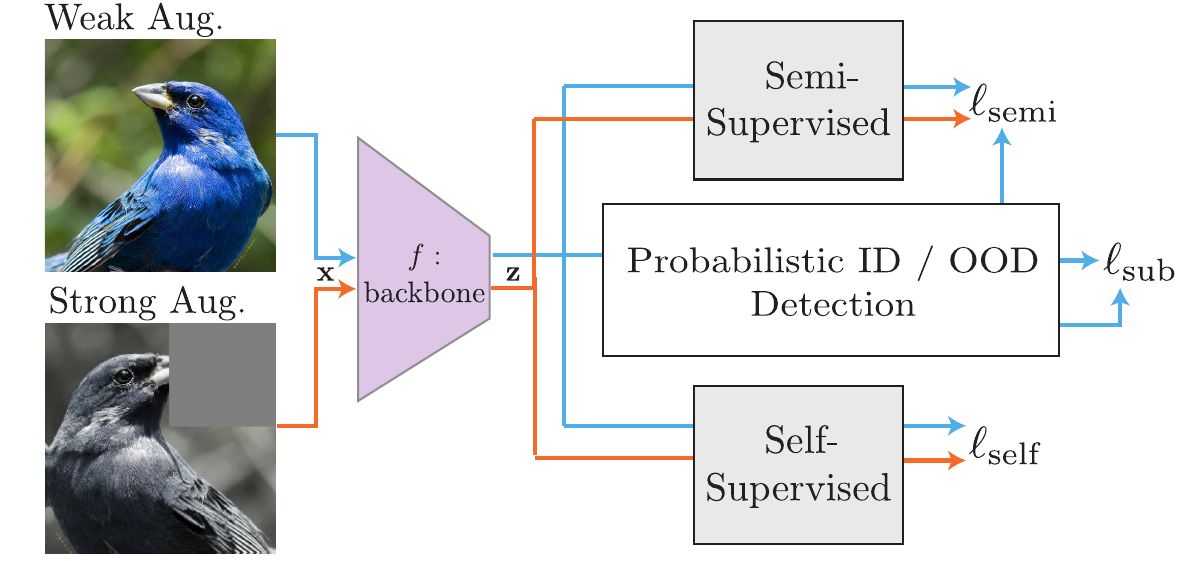}
    \unskip\ \vrule\
    \raisebox{0.6cm}{\includegraphics[width=0.44\linewidth]{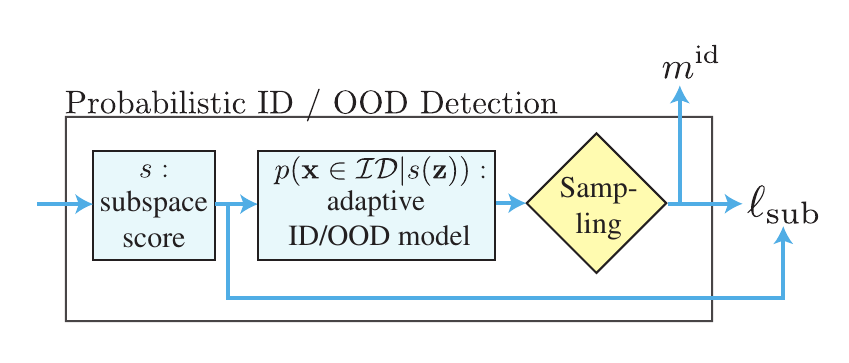}}
    \caption{\textbf{Left}: The flow of unlabeled data in ProSub. \textbf{Right}: Details for ID/OOD detection, the main contribution of ProSub.}
    \label{fig:method-graph}
\end{figure*}

The proposed method, ProSub, can be summarized as handling unlabeled data through three main components, as shown in \cref{fig:method-graph}. First, we adopt self-supervision as proposed in \cite{wallin2023improving} to enable learning feature representations from all unlabeled data, both ID and OOD (see \cref{sec:self-supervision}). Second, we use a similar pseudo-labeling strategy as \cite{sohn2020fixmatch} to assign unlabeled data to ID classes in a cross-entropy loss (see \cref{sec:pseudo-labeling}). However, to avoid assigning pseudo-labels to OOD data, we want to exclude these data here. To this end, we propose a component for probabilistic ID/OOD detection, which is also the main contribution of ProSub. This component samples binary labels for unlabeled data from a posterior distribution, marking them as ID or OOD. These labels are used to disable pseudo-labeling for data marked as OOD.

The ID/OOD module of ProSub consists of first predicting the subspace score for each sample given its features, $s(\mathbf{z})$ (see \cref{sec:subspace-score}). Subsequently, we use estimates of the conditional distributions of scores given ID or OOD data, $p_\text{id}(s)$ and $p_\text{ood}(s)$, which by Bayes' theorem enable probabilistic predictions of samples being ID or OOD as
\begin{equation} \label{eq:p-id}
    p(\mathbf{x} \in \mathcal{ID} | s(\mathbf{z})) = \frac{\pi p_\text{id}(s(\mathbf{z}))}{\pi p_\text{id}(s(\mathbf{z})) + (1 - \pi) p_\text{ood}(s(\mathbf{z}))},
\end{equation}
where $\pi$ is the proportion of ID data in the marginal distribution of both ID and OOD data. The set of ID data is denoted by $\mathcal{ID}$. The predicted probability is then used to sample the binary ID/OOD labels. Finally, the performance of the subspace score is enhanced through a subspace loss (see \cref{sec:l-sub}). This loss utilizes the binary labels to further separate the distributions of scores for ID and OOD. We now move on to describe the parts of ProSub in more detail.

\subsection{Proposing the Subspace Score} \label{sec:subspace-score}

Deep neural networks trained for classification in a fully supervised, closed-set setting with cross-entropy loss have been shown to follow the principles of neural collapse in their terminal training stage (when full training accuracy is reached) \cite{papyan2020prevalence}. This (empirical) phenomenon is defined by a set of characteristics exhibited by features within the output of the penultimate network layer. For our purpose, the key property of neural collapse is the convergence of features from training data converge towards class means, $\mathbf{c}_1, \mathbf{c}_2, \ldots, \mathbf{c}_C$ for $C$ classes.

While OSSL differs from the fully-supervised setting discussed in \cite{papyan2020prevalence}, we observe that models trained using SSL quickly overfit the small labeled training set, suggesting that the features of labeled data may follow the principles of neural collapse. Assuming that features of labeled data collapse to the class means, one can try to distinguish ID data from OOD data by measuring the distance to the set of feature means, where a large distance indicates data being OOD. We compared several such measures (see \cref{sec:subspace-alternatives}). We find that, in combination with a cosine-based self-supervision, the best-performing method is to measure the angle between the space spanned the class means, see \cref{fig:firstpagefig}.

Specifically, we first compute the ID subspace, $W_\text{id}$, as the space spanned by $C$ class means:
\begin{equation}
    W_\text{id} = \text{span} \left( \{ \mathbf{c}_1, \mathbf{c}_2, \dots, \mathbf{c}_C \} \right),
\end{equation}
where $\mathbf{c}_c \in \R^D$, $c=1, \dots, C$ are the class means associated with each class, calculated from labeled data. Then, given a predicted feature vector of a test sample $\mathbf{z}$, we want to get the angle between the test vector and $W_\text{id}$. This is achieved by first finding an orthonormal basis of $W_\text{id}$ through QR decomposition \cite{golub2013matrix} of the matrix whose columns vectors are $\mathbf{c}_c$ for $c = 1, \dots, C$: $\mathbf{C} \in \R^{D \times C}$. The columns of $\mathbf{Q}$ from the decomposition $\mathbf{Q}\mathbf{R} = \mathbf{C}$ then form the orthonormal basis on $W_\text{id}$ and the projection of $\mathbf{z}$ on $W_\text{id}$ is $\text{proj}_{W_\text{id}}(\mathbf{z}) = \mathbf{Q} \mathbf{Q}^T \mathbf{z}$.The subspace score that we propose, $s(\cdot)$, is the cosine of the angle between $\mathbf{z}$ and $W_\text{id}$:  
\begin{equation} \label{eq:betamodel-id-score}
    s(\mathbf{z}) = \frac{  \text{proj}_{W_\text{id}}(\mathbf{z}) \cdot \mathbf{z}}{\| \text{proj}_{W_\text{id}}(\mathbf{z}) \| \| \mathbf{z} \|}.
\end{equation}
Empirical results show that, for ID data, $\mathbf{z}$ will have a small angle to $W_\text{id}$ and $s(\mathbf{z})$ will be close to one, whereas for OOD data, $s(\mathbf{z})$ will be closer to zero.

The class means of $\mathbf{C}$ are obtained by evaluating the exponential moving averages (EMA) of features for labeled data. For $c=1, \dots, C$, in each training step, we get for every batch containing class $c$ samples
\begin{equation} \label{eq:betamodel-prototype-updates}
\begin{gathered}
    \mathbf{c}_c \leftarrow \lambda \mathbf{c}_c + (1-\lambda) \frac{\sum_{i=1}^B \mathds{1}\{ y_i = c \} \mathbf{z}_i^l}{\sum_{i=1}^B\mathds{1}\{ y_i = c \}},
\end{gathered}
\end{equation}
where $\lambda$ is the momentum for the EMA, $\mathbf{z}_i^l$ are the predicted feature vectors for labeled samples in the batch, $y_i$ are the labels for samples in the batch, $B$ is the size of the labeled batch, and $\mathds{1}\{\cdot\}$ is the indicator function.

\subsection{Estimating a Probabilistic Model}

To enable probabilistic predictions of data being ID or OOD, as specified in \eqref{eq:p-id}, we need models for $p_\text{id}(s)$ and $p_\text{ood}(s)$. The \textbf{Beta distribution} \cite{johnson1995continuous} is a distribution with the desired properties: support on $[0,1]$, a flexible shape, and closed-form estimation methods. Additionally, we empirically find that our data fit the Beta distribution well (see \cref{sec:results-densities}). The Beta distribution has two positive parameters, $\alpha$ and $\beta$, that we need to estimate for both $p_\text{id}(s)$ and $p_\text{ood}(s)$. However, when we observe score samples, we usually do not know if the data are ID or OOD, making the estimation challenging.

One approach for estimating models depending on hidden variables is to use MLE through the iterative EM algorithm \cite{dempster1977maximum}, with the component association (data being ID or OOD) representing the hidden variable for our case. The EM algorithm involves alternation between an E-step and an M-step until convergence. In the E-step, we compute probabilities for component associations (weights) given our current estimates of $\alpha$ and $\beta$ for ID and OOD. The M-step uses these probabilities in a weighted MLE for each separate component to improve the estimate. However, the Beta distribution has no closed-form expression for MLE, necessitating expensive numerical solutions in the M-step \cite{beckman1978maximum}.

To simplify the M-step, an alternative is the ad-hoc replacement of weighted MLE with the method of moments estimate, which has a closed-form solution for the Beta distribution. This approach is introduced in \cite{schroder2017hybrid} as the \textbf{iterated method of moments} (IMM). Although no longer maximizing the overall likelihood, IMM works well in practice. Specifically, IMM replaces the M-step with an MM-step that first involves computing weighted sample moments:
\begin{equation} \label{eq:weighted-estimations}
        \tilde{\mu} = \frac{1}{\sum_{i=1}^n w_i} \sum_{i=1}^n w_i s_i, \qquad
        \tilde{\sigma}^2 = \frac{1}{\sum_{i=1}^n w_i} \sum_{i=1}^n w_i (s_i - \tilde{\mu})^2,
\end{equation}
where $s_i$ are the score samples and $w_i$ are the corresponding weights from the E-step. These moments are then used to estimate $\alpha$ and $\beta$ through the method of moments as
\begin{equation} \label{eq:mm-beta}
    \alpha = \tilde{\mu} \left( \frac{\tilde{\mu}(1 - \tilde{\mu})}{\tilde{\sigma}^2} - 1 \right), \qquad
    \beta = (1 -  \tilde{\mu}) \left( \frac{\tilde{\mu}(1 - \tilde{\mu})}{\tilde{\sigma}^2} - 1 \right).
\end{equation}

Another challenge arising for OSSL is that we need accurate estimates of $p_\text{id}(s)$ and $p_\text{ood}(s)$ during the full training duration but our network is continuously changing. Estimates using the full dataset until convergence in each training step are impractically expensive. One remedy close at hand is to carry out the estimation at pre-defined intervals and assume that the estimated parameters are valid until the next estimation. However, network parameters during training can be noisy and there is no guarantee that the training steps we use for the estimations yield models that are accurate for the upcoming interval.

To adapt the estimation for the OSSL setting, we propose a batch version of IMM in which we perform one E-step and one MM-step in each training step, using only the data of the current batch. The parameters of the conditionals, $\alpha_\text{id}$, $\beta_\text{id}$, $\alpha_\text{ood}$, and $\beta_\text{ood}$, are updated as an EMA of the batch estimates. Additionally, since each batch contains known ID data points, the labeled data, we include these with weights equal to 1.0 in the estimation of $\alpha_\text{id}$ and $\beta_\text{id}$. This approach is outlined in \cref{alg:em-algorithm}. We find empirically that this procedure produces accurate estimates for the full training duration (see \cref{sec:results-densities}).

\subsection{Enhancing OOD Detection with a Subspace Loss} \label{sec:l-sub}

To improve performance for ID/OOD classification, we include a subspace loss to increase $s$ for ID data and decrease $s$ for OOD data. In each training step for unlabeled data, we calculate probabilities of samples being ID using \eqref{eq:p-id}: $p^\text{id}_i$, $i=1,\dots, \mu B$, where $\mu B$ is the unlabeled batch size. Given these probabilities, we randomly sample an ID mask as
\begin{equation}\label{eq:id-sampling}
        m^\text{id}_i = \mathds{1}\{ p^\text{id}_i \geq X_i \}, \; X_i \sim U(0,1), \;
        \text{for} \; i=1,\dots,\mu B.
\end{equation}
Consequently, we get a corresponding OOD mask $m^\text{ood}_i = 1 - m^\text{id}_i$. For data sampled as ID, we encourage the model to increase $s$, and for data sampled as OOD, we encourage the model to decrease $s$. The resulting loss is
\begin{equation} \label{eq:betamodel-align}
    \ell_\text{sub} = \frac{1}{\mu B} \sum_{i=1}^{\mu B} (m_i^\text{ood} - m_i^\text{id}) s(\mathbf{z}_i),
\end{equation}
where $\mathbf{z}_i$ are the features for unlabeled data. The class means $\mathbf{C}$, used to calculate $s$, are considered constant when computing gradients w.r.t. $l_\text{sub}$. \cref{sec:subspace-alternatives} discusses an alternative $l_\text{sub}$ that uses $p_i^\text{id}$ directly instead of the random mask.

\subsection{Pseudo-labeling} \label{sec:pseudo-labeling}

We adopt a similar pseudo-labeling strategy as FixMatch \cite{sohn2020fixmatch}. However, in addition to requiring predictions to exceed a confidence threshold, $\tau$, we also require data to be sampled as ID, following \eqref{eq:id-sampling}. The resulting pseudo-labeling loss is
\begin{equation} \label{eq:betamodel-pl-loss}
\begin{gathered}
    \ell_\text{semi} = \frac{1}{\mu B} \sum_{i=1}^{\mu B} \mathds{1}\{ \max_{y'} p_{\bm{\theta}}(y' | \mathbf{x}_i) > \tau \land m^\text{id}_i = 1 \} \\
    \times H \left( \argmax_{y'} [ p_{\bm{\theta}} (y' | \mathbf{x}_i) ], p_{\bm{\theta}} (y | \tilde{\mathbf{x}}_i) \right),
\end{gathered}
\end{equation}
where $\land$ denotes the logical \emph{and} operation, $\mathbf{x}_i$ are (weakly augmented) unlabeled samples, $\tilde{\mathbf{x}}_i$ are strongly augmented unlabeled samples, and $H(\cdot, \cdot)$ is the cross entropy. When computing gradients with respect to $\ell_\text{semi}$, predictions on weakly augmented data, $\mathbf{x}_i$, are considered constant.

\subsection{Self-supervision} \label{sec:self-supervision}

Following \cite{wallin2023improving}, to enable learning from all unlabeled data, both ID and OOD, we include a cosine-based self-supervision, defined as
\begin{equation} \label{eq:betamodel-self-supervision}
    \ell_\text{self} = - \frac{1}{\mu B} \sum_{i=1}^{\mu B} \frac{ h(\tilde{\mathbf{z}}_i) \cdot \mathbf{z}_i }{ \| h( \tilde{\mathbf{z}}_i ) \| \cdot \| \mathbf{z}_i \| },
\end{equation}
where $h(\cdot)$ is a trainable linear transformation, $\tilde{\mathbf{z}}_i$ and $\mathbf{z}_i$ are predicted feature vectors for strongly augmented and weakly augmented unlabeled samples, respectively. Again, the predictions on weakly augmented data are considered constant when computing the gradients w.r.t. this loss.

\subsection{Final Training Objective}

In line with the established convention in SSL \cite{laine2016temporal, tarvainen2017mean, sohn2020fixmatch}, we use a standard supervised cross-entropy loss on labeled data, given by
\begin{equation} \label{eq:betamodel-ll}
    \ell_\text{sup} = \frac{1}{B} \sum_{i=1}^B H(y_i, p_{\bm{\theta}} (y | \mathbf{x}_i^l)),
\end{equation}
were $y_i$ is the label for sample $i$ and $\mathbf{x}_i^l$ are the labeled samples. As another prevalent component in SSL \cite{berthelot2019remixmatch, zhang2021flexmatch}, we include $l^2$-regularization on the model parameters $\bm{\theta}$, given by
$\ell_\text{reg} = \frac{1}{2} \| \bm{\theta} \|^2$.

Putting it all together, our final training objective is a weighted sum:
\begin{equation} \label{eq:full-loss}
    \ell = \ell_\text{sup} + w_\text{semi} \ell_\text{semi} + w_\text{self} \ell_\text{self} + w_\text{sub} \ell_\text{sub} + w_\text{reg} \ell_\text{reg},
\end{equation}
where $w_\text{semi}$, $w_\text{self}$, $w_\text{sub}$, and $w_\text{reg}$ are scalars controlling the importance of each term. Since $\ell_\text{semi}$, similarly to $\ell_\text{sup}$, is a cross-entropy, $w_\text{semi} = 1.0$ is typically a good choice. We empirically find $w_\text{sub}=1.0$ effective. The self-supervision $w_\text{self}$ benefits from some tuning. Suitable values for $w_\text{reg}$ can be found in the literature. See \cref{sec:implementation} and the supplementary material for more details on these weights.

\begin{figure}[!t]
\begin{minipage}[]{0.5\linewidth}
\RestyleAlgo{boxruled}
\begin{algorithm}[H]
    \scriptsize
    \caption{Batch IMM for estimating $p_\text{id}(s)$ and $p_\text{ood}(s)$} \label{alg:em-algorithm}
    \DontPrintSemicolon
    \input{algorithms/em-algorithm}
\end{algorithm}    
\end{minipage} %
\begin{minipage}[]{0.5\linewidth}
\RestyleAlgo{boxruled}
\begin{algorithm}[H]
    \scriptsize
    \caption{Training step for ProSub. \vphantom{$p_\text{ood}(s)$}} \label{alg:beta-model-training-step}
    \DontPrintSemicolon
    \input{algorithms/beta-model-training-step}
\end{algorithm}    
\end{minipage}
\end{figure}

\subsection{Optimization and Data Augmentation}

Following many existing SSL works, we use SGD with Nesterov momentum and a cosine decay for the learning rate \cite{sohn2020fixmatch, zhang2021flexmatch, saito2021openmatch}. We use a warm-up phase with a constant learning rate to allow scores and estimates to settle before applying all losses. Specifically, The learning rate, $\eta$, follows the schedule given by
\begin{equation}
    \label{eq:betamodel-lr}
    \eta(k) = 
    \begin{cases}
    \eta_0 \quad &\text{for} \quad k < K_p \\
    \eta_0 \cos \left( \gamma \frac{\pi (k-K_p)}{2 (K-K_p)} \right) \quad &\text{otherwise}
    \end{cases},
\end{equation}
where $\eta_0$ denotes the initial learning rate, $K_p$ and $K$ are the number of warm-up steps and the total number of training steps, respectively, and $k$ is the current training step. The decay rate is controlled by $\gamma$.

For data augmentation, we follow the strategy of FixMatch \cite{sohn2020fixmatch}, using stochastic flip and translation for weak augmentation, and two operations from Randaugment \cite{cubuk2020randaugment} followed by Cutout \cite{devries2017improved} for strong augmentations.

Training steps of ProSub are detailed in \cref{alg:beta-model-training-step}. In the warm-up phase, we use $\ell_\text{sup}$, $\ell_\text{self}$, and $\ell_\text{reg}$. In the subsequent training phase, the pseudo-labeling loss, $\ell_\text{semi}$, and the subspace loss $\ell_\text{sub}$ are added to the training objective. In \cref{alg:beta-model-training-step}, we denote the backbone model $f(\cdot)$, that predicts features given input data, the classification model that predicts the class distribution given features is denoted $g(\cdot)$, finally the projection head used in \eqref{eq:betamodel-self-supervision} is denoted $h(\cdot)$.

\section{Experiments and Results}
\label{sec:experiments}

We follow the evaluation procedure of \cite{wallin2023improving}, using datasets CIFAR-10/100 \cite{krizhevsky2009learning} as ID with (the other) CIFAR-10/100 as OOD. Following \cite{saito2021openmatch}, we evaluate ImageNet30 with 2,600 labels using 20 classes as ID and 10 classes as OOD. We also evaluate Tiny ImageNet \cite{le2015tiny} (5,000 labels) with 100 classes as ID and 100 classes as OOD, and ImageNet100 \cite{imagenet100} (5,000 labels) with 50 classes as ID and 50 classes as OOD. The model is evaluated in terms of closed-set accuracy and AUROC for ID/OOD classification on test sets at the end of training (a few runs for OpenMatch use early stopping with validation data to avoid collapse). Baseline results are taken from \cite{wallin2023improving, saito2021openmatch} when available. Evaluations new for this work are reported as mean and std over three runs using EMA of model parameters.

We compare to OSSL methods that prioritize both closed-set accuracy and OOD detection. We have focused on works that are published in peer-reviewed publications with released code: MTCF \cite{yu2020multi}, T2T \cite{huang2021trash}, OpenMatch \cite{saito2021openmatch}, IOMatch \cite{li2023iomatch}, and SeFOSS \cite{wallin2023improving}. Recent works such as \cite{fan2023ssb, ma2023rethinking} are interesting but unfortunately do not have available code at the time of writing, making fair comparisons difficult. We include the (closed-set) SSL baseline FixMatch \cite{sohn2020fixmatch} and a supervised model using only the labeled data; these use the energy score \cite{li2020energy} for OOD detection. Results are shown in \cref{tab:betamodel-results}.

\begin{table*}[!t]
    \scriptsize
    \centering
    \caption{Closed-set accuracy (top rows) and AUROC for ID/OOD classification (bottom rows). Dagger$^\dagger$ marks using labeled validation data for early stopping. \textbf{Boldface} denotes best accuracies among OSSL methods and \underline{underline} denotes best AUROCs.} %
    \setlength{\tabcolsep}{1.0pt}
    \resizebox{\linewidth}{!}{%
    \begin{tabular}{@{}r c c c c c c c@{}}
\toprule

& \multicolumn{2}{c}{ID: CIFAR-10} & \multicolumn{2}{c}{ID: CIFAR-100}
& & & \\
& \multicolumn{2}{c}{OOD: CIFAR-100} & \multicolumn{2}{c}{OOD: CIFAR-10}
& \multirow{2}{*}{IN 20/10} & \multirow{2}{*}{IN 50/50} & \multirow{2}{*}{TIN 100/100}\\ \cmidrule(lr){2-3} \cmidrule(lr){4-5}
& 1,000 lab. & 4,000 lab. 
& 2,500 lab. & 10,000 lab. & & & \\ \midrule

\multirow{2}{*}{Only labeled} 
& 54.51\spm{1.82} & 75.57\spm{2.88} 
& 34.62\spm{1.43} & 59.12\spm{0.91}
& 63.15\spm{2.95} & 38.17\spm{0.83} & 38.12\spm{1.20}\\
& 0.62\spm{0.01} & 0.74\spm{0.02} 
& 0.61\spm{0.01} & 0.71\spm{0.01}
& 0.71\spm{0.02} & 0.61\spm{0.01} & 0.61\spm{0.00}\\ 
\midrule 

\multirow{2}{*}{FixMatch \cite{sohn2020fixmatch}} 
& 92.70\spm{0.14} & 94.07\spm{0.15} 
& 71.95\spm{0.49} & 77.72\spm{0.32}
& 94.11\spm{0.15} & 69.81\spm{0.44} & 59.86\spm{0.18}\\
& 0.66\spm{0.00} & 0.69\spm{0.01} 
& 0.46\spm{0.01} & 0.51\spm{0.01}
& 0.52\spm{0.02} & 0.48\spm{0.01} & 0.57\spm{0.00}\\ 
\midrule %

\multirow{2}{*}{MTCF \cite{yu2020multi}} 
& 82.96\spm{1.08} & 89.87\spm{0.21} 
& 40.46\spm{1.49} & 62.88\spm{0.92}
& 86.40\spm{0.70} & 50.65\spm{0.80} & 39.55\spm{0.23} \\
& 0.81\spm{0.00} & 0.84\spm{0.00} 
& 0.82\spm{0.01} & 0.80\spm{0.01}
& 0.94\spm{0.00} & 0.82\spm{0.00} & 0.59\spm{0.00}\\ 
\cdashlinelr{2-8}

\multirow{2}{*}{T2T \cite{huang2021trash}} & 
86.99\spm{1.09} & 86.11\spm{1.91}
& 38.30\spm{9.72} & 62.02\spm{3.73}
& 89.81\spm{0.35} & 54.17\spm{5.81} & 45.70\spm{0.71}\\
& 0.57\spm{0.02} & 0.57\spm{0.04}
& 0.63\spm{0.08} & 0.59\spm{0.08}
& 0.80\spm{0.01} & 0.64\spm{0.04} & 0.61\spm{0.00}\\ 
\cdashlinelr{2-8}

\multirow{2}{*}{OpenMatch \cite{saito2021openmatch}} 
& 92.20\spm{0.15} & \textbf{94.82}\spm{0.21}
& $^\dagger$63.33\spm{0.86} & $^\dagger$75.89\spm{0.23} %
& 89.60\spm{1.00} & 58.23\spm{0.15} & 53.82\spm{0.11}\\
& \underline{0.93}\spm{0.00} & \underline{0.96}\spm{0.00}
& $^\dagger$0.86\spm{0.01} & $^\dagger$0.92\spm{0.01} %
& 0.96\spm{0.00} & 0.82\spm{0.00} & 0.66\spm{0.00}\\ 
\cdashlinelr{2-8}

\multirow{2}{*}{IOMatch \cite{li2023iomatch}} 
& 91.77\spm{0.28} & 93.34\spm{0.05}
& 68.89\spm{0.18} & 75.82\spm{0.28}
& 87.52\spm{1.18} & 47.03\spm{1.13} & 57.77\spm{0.37} \\
& 0.69\spm{0.01} & 0.74\spm{0.00}
& 0.56\spm{0.01} & 0.58\spm{0.00}
& 0.80\spm{0.02} & 0.63\spm{0.01} & 0.62\spm{0.00} \\
\cdashlinelr{2-8}

\multirow{2}{*}{SeFOSS \cite{wallin2023improving} } 
& 91.49\spm{0.16} & 93.73\spm{0.27}
& 68.48\spm{0.26} & 77.63\spm{0.21}
& 92.53\spm{0.10} & 69.20\spm{0.44} & 59.18\spm{0.50}\\
& 0.90\spm{0.01} & 0.92\spm{0.00} 
& 0.79\spm{0.01} & 0.83\spm{0.00}
& 0.97\spm{0.00} & 0.80\spm{0.05} & 0.61\spm{0.00}\\
\cdashlinelr{2-8}

\multirow{2}{*}{ProSub (ours)} 
& \textbf{92.81}\spm{0.60} & 94.50\spm{0.05} 
& \textbf{74.16}\spm{0.49} & \textbf{79.59}\spm{0.37}
& \textbf{93.37}\spm{0.41} & \textbf{71.15}\spm{0.80} & \textbf{60.92}\spm{0.32}\\
& 0.92\spm{0.00} & 0.93\spm{0.00} 
& \underline{0.97}\spm{0.00} & \underline{0.98}\spm{0.00}
& \underline{0.98}\spm{0.00} & \underline{0.96}\spm{0.00} & \underline{0.72}\spm{0.00}\\ 

\bottomrule
\end{tabular}
    }
    \label{tab:betamodel-results} %
\end{table*}

ProSub yields the best results for both closed-set accuracy and OOD detection in most scenarios. Noteworthy are the large improvements in AUROC when CIFAR-100 is ID, and on ImageNet50/50. OpenMatch \cite{saito2021openmatch} performs slightly better than ProSub when CIFAR-10 is ID. An explanation for this is that the cosine-based self-supervision is less effective for CIFAR-10, since \cite{wallin2023improving, wallin2022doublematch} report comparably worse results for CIFAR-10. We also note that ProSub outperforms FixMatch \cite{sohn2020fixmatch} in closed-set accuracy on many scenarios, even though FixMatch is a method that does not consider ID/OOD classification.

\subsection{Implementation Details} \label{sec:implementation}

We use architectures WRN-28-2 \cite{zagoruyko2016wide} when CIFAR-10 is ID, WRN-28-8 when CIFAR-100 is ID, WRN-28-4 for TIN, and ResNet18 \cite{he2016deep} for IN20/10 and IN50/50. For the subspace loss, we use $w_\text{sub} = 1.0$. We use $w_\text{self} = 10$ when CIFAR-10 is ID, $w_\text{self} = 15$ when CIFAR-100 is ID,  $w_\text{self} = 20$ for IN20/10, $w_\text{self} = 50$ for TIN, and $w_\text{self}=40$ for IN50/50. We use $K_p=5\cdot10^4$ and $K=2^{19}$, except for IN20/10 and IN50/50 where we use $K_p=3 \cdot 10^4$ and $K=10^5$. Other hyperparameters are the same as in \cite{wallin2023improving}.
We use $\pi$ matching the actual unlabeled distributions and show in the supplementary material that this choice is not critical to our performance. In addition, we include an extended discussion on hyperparameter selection and limitations. 
For T2T, SeFOSS, OpenMatch, and IOMatch, we use the official implementations with original hyperparameters (except $w_\text{self}$ for SeFOSS which follows the values specified here).

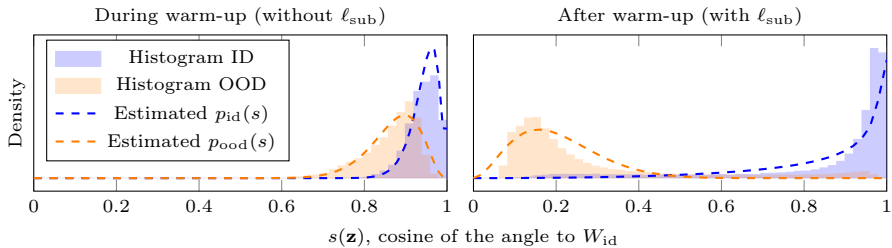
\begin{figure}[!t]
    \centering
    \begin{tikzpicture}

    \newcommand{\myheight}{0.35\textwidth}

    \begin{groupplot}[
        group style={
            group size=2 by 1,
            x descriptions at=edge bottom,             
            horizontal sep=10pt,
        },
    ]
    \nextgroupplot[
        title style={font=\scriptsize},
        label style={font=\scriptsize},
        title style={yshift=-1.5ex},        
        title={During warm-up (without $\ell_\text{sub}$)},
        width=0.58\columnwidth,
        height=0.3\textwidth,
        ylabel={Density},
        ytick=\empty,
        ticklabel style = {font=\scriptsize},
        xmin=0, xmax=1,
        legend pos=north west,
        legend style={font=\scriptsize},
    ]

        \addplot [name path=ZERO,draw=none, forget plot] coordinates {(0,0) (1,0)};
        
        \addplot[color=blue, draw=none, name path=ID, forget plot]
        table[col sep=comma, x=x, y=y] {figure-data/id_hist_prealign.csv};
        \addplot [blue, opacity=0.2] fill between [of=ID and ZERO];        
        
        \addplot[color=orange, draw=none, name path=OOD, forget plot]
        table[col sep=comma, x=x, y=y] {figure-data/ood_hist_prealign.csv};
        \addplot [orange, opacity=0.2] fill between [of=OOD and ZERO]; 
        
        \addplot[color=blue, thick, dashed]
        table[col sep=comma, x=x, y=y] {figure-data/beta_id_prealign.csv};

        \addplot[color=orange, thick, dashed]
        table[col sep=comma, x=x, y=y] {figure-data/beta_ood_prealign.csv};
        
        \legend{Histogram ID, Histogram OOD, Estimated $p_\text{id}(s)$, Estimated $p_\text{ood}(s)$}

    \nextgroupplot[
        title style={font=\scriptsize},
        label style={font=\scriptsize},  
        title style={yshift=-1.5ex},
        title={After warm-up (with $\ell_\text{sub}$)},
        width=0.58\columnwidth,
        height=0.3\textwidth,
        ylabel=\empty,
        ytick=\empty,
        ticklabel style = {font=\scriptsize},
        xlabel={$s(\mathbf{z})$, cosine of the angle to $W_\text{id}$},
        every axis x label/.append style={at=(ticklabel cs:0.0)},
        xmin=0, xmax=1,
        legend pos=south east,
    ]

        \addplot [name path=ZERO,draw=none] coordinates {(0,0) (1,0)};

        \addplot[color=black, draw=none, name path=ID]
        table[col sep=comma, x=x, y=y] {figure-data/id_hist.csv};
        \addplot [blue, opacity=0.2] fill between [of=ID and ZERO];
        
        \addplot[color=black, draw=none, name path=OOD]
        table[col sep=comma, x=x, y=y] {figure-data/ood_hist.csv}; 
        \addplot [orange, opacity=0.2] fill between [of=OOD and ZERO];        
        
        \addplot[color=blue, thick, dashed]
        table[col sep=comma, x=x, y=y] {figure-data/beta_id.csv};

        \addplot[color=orange, thick, dashed]
        table[col sep=comma, x=x, y=y] {figure-data/beta_ood.csv};    

    \end{groupplot}
    
\end{tikzpicture}
    \caption{Results from our estimation approach as specified in \cref{alg:em-algorithm} both from the warm-up stage and the subsequent training stage (with $\ell_\text{sub}$ applied).}
    \label{fig:em-estimation}
\end{figure}

\subsection{Analyzing Density Estimation and $\ell_\text{sub}$} \label{sec:results-densities}

To assess the accuracy of our estimates of $p_\text{id}(s)$ and $p_\text{ood}(s)$, we compare the empirical distributions of scores given ID and OOD data with the estimates obtained from our IMM approach as specified in \cref{alg:em-algorithm}. This is done in the warm-up phase (before $\ell_\text{sub}$ is applied) and after the warm-up stage. Specifically, we use CIFAR-100 (2,500 labels) as ID with CIFAR-10 as OOD and evaluate at training steps 40,000 and 80,000. \cref{fig:em-estimation} shows that our IMM approach successfully estimates the distributions of scores for ID and OOD data both when there is a large overlap and when they are separated. Note that the estimation algorithm only has access to the marginal of the empirical distributions, not the plotted conditionals. \cref{fig:em-estimation} also highlights the effect of $\ell_\text{sub}$: in the warm-up phase, there is overlap between scores of ID and OOD data, application of $\ell_\text{sub}$ then successfully creates a separation between the two conditionals.

\subsection{Ablation: Self-supervision Enables the Subspace Score} \label{sec:enables-subspace}

To compare our proposed score with baselines, we evaluate the AUROC using different scores for ID/OOD classification in ProSub. This evaluation is done at the end of the warm-up phase before any loss that directly alters these scores has been used. We use CIFAR-100 (2,500 labels) as ID with CIFAR-10 as OOD, and ImageNet50/50 with 5,000 labels. Furthermore, we compare to a fully supervised model trained using labels for \emph{all} ID data (but not exposed to OOD data). Our evaluations include the OOD detection baselines maximum softmax probability (MSP) \cite{hendrycks2016baseline}, the energy-based score \cite{li2020energy}, and the max logit score \cite{vaze2021open}. The results are shown in \cref{tab:betamodel-scores}. The subspace score outperforms the baselines for OOD detection in ProSub. However, for the fully supervised model, we see the opposite relation. This indicates that the training signal from unlabeled data (ID and OOD) through the cosine-based self-supervision specified in \eqref{eq:betamodel-self-supervision} is key for enabling the strong performance of the subspace score in OSSL. Note that each column of \cref{tab:betamodel-scores} uses one model (with different scores), so closed-set accuracies within each column are equal (shown in parenthesis).

\begin{table}[!t]
    \scriptsize
    \centering
    \caption{AUROCs for ID/OOD classification using different scores for ProSub (without $\ell_\text{semi}$ and $\ell_\text{self}$) and a fully supervised model using labels for \emph{all} ID data.}    
    \begin{tabular}{c c c c c}
\toprule
& \multicolumn{2}{c}{ID: CIFAR-100, OOD: CIFAR-10} & \multicolumn{2}{c}{ImageNet 50/50} \\ \cmidrule(lr){2-3}\cmidrule(lr){4-5}
& ProSub (62\%) & Fully supervised (79\%) & ProSub (64\%) & Fully supervised (72\%) \\ \midrule
MSP & 0.63 & 0.79 & 0.77 & 0.77 \\
Energy & 0.65 & 0.81 & 0.80 & 0.79 \\
Max logit & 0.65 & 0.81 & 0.80 & 0.79 \\
$s$ (ours) & \underline{0.92} & 0.73 & \underline{0.93} & 0.58 \\
\bottomrule
\end{tabular}%
    \label{tab:betamodel-scores}
\end{table}

\subsection{Ablation: Alternative Designs for the Subspace Score} \label{sec:subspace-alternatives}

In ProSub, we use the subspace score, $s(\cdot)$, for ID/OOD classification, as specified in \eqref{eq:betamodel-id-score}. This score relies on the angles between features, $\mathbf{z}$, and $W_\text{id}$, the space spanned by the class means, $\mathbf{c}_1, \mathbf{c}_2, \ldots, \mathbf{c}_C$. There are alternative ways to evaluate the distance between the set of class means and features. We investigate three of these and compare how they perform to our subspace score: 1) the negated minimum Euclidean distance to $\mathbf{c}_c$: $- \min_c \|\textbf{z} - \textbf{c}_c \|$, 2) the negated Euclidean distance to $W_\text{id}$: $- \| \mathbf{z} - \text{proj}_{W_\text{id}}(\mathbf{z}) \|$, and 3) the maximum cosine similarity to $\mathbf{c}_c$: $\max_c \textbf{z} \cdot \textbf{c}_c / (\| \textbf{z} \| \| \textbf{c}_c \|)$. Note the similarity of 2) to Vim \cite{wang2022vim}.

\Cref{tab:subspace-alternatives} shows AUROC for OOD detection at the end of warm-up in ProSub using $s$ and these alternative scores. We evaluate at the end of the warm-up phase to avoid the subspace loss, $\ell_\text{sub}$ (see \eqref{eq:betamodel-align}), influencing the results. Similarly to \cref{tab:betamodel-scores}, each column in \cref{tab:subspace-alternatives} uses one model evaluated with different scores; the closed-set accuracies for these models are shown in the bottom row. We use CIFAR-100 (2,500 labels) with CIFAR-10 as OOD, TIN100/100, and ImageNet50/50. The subspace score, $s$, gives the best results for all datasets. The second best score is the max similarity, which is also cosine-based, indicating that a cosine-based self-supervision facilitates a cosine-based ID/OOD score.

A hypothesis for why the subspace score performs better than the max similarity is that $s$ is \emph{class agnostic}, \ie, the model can identify a sample as ID but be uncertain about the specific class. A sample can, \eg, be placed between two class means on $W_\text{id}$, yielding a large $s$, but not a large value for the max similarity. Empirically, this is supported by \cref{fig:tosub-vs-insub} showing that the spread of angles for ID data to $W_\text{id}$ is smaller than the spread of angles \emph{within} $W_\text{id}$ to the closest class-mean. \cref{fig:tosub-vs-insub} shows results using CIFAR-100 (2,500 labels) as ID and CIFAR-10 as OOD at the end of the warm-up phase. The architecture is WRN-28-8 which gives a feature space of dimension 512 whereas the (maximum) dimension of $W_\text{id}$ corresponds to the number of classes, which is 100.

The next advantage of $s$ compared to the alternatives is that it is well modeled by the Beta distribution and that the mixture of scores for ID and OOD can be estimated through our iterative algorithm, see \cref{sec:results-densities}. While there might exist distributions that successfully model the other scores presented in this section, and corresponding estimation procedures, this is no guarantee. For example, we quickly see from \cref{fig:tosub-vs-insub} that the angles within $W_\text{id}$ do not follow the shape of the Beta distribution, and it is not clear that we accurately can represent this distribution with a single parametric model.

\begin{figure}[t]
\begin{minipage}{0.48\linewidth}
    \scriptsize
    \centering
    \captionof{table}{AUROCs for alternatives to $s$.}%
    \label{tab:subspace-alternatives}
    \begin{tabular}{c c c c}
\toprule
& CIFAR-100 & IN & TIN \\ 
& CIFAR-10 & 50/50 & 100/100 \\ \cmidrule{2-4}
1) Min dist. & 0.70 & 0.63 & 0.52 \\
2) Residual dist. & 0.88 & 0.63 & 0.59 \\
3) Max sim. & 0.87 & 0.91 & 0.67 \\
$s$ (ours) & \underline{0.92} & \underline{0.93} & \underline{0.68} \\ \cdashlinelr{1-4}
Acc. & 62\% & 64\% & 54\% \\
\bottomrule
\end{tabular}
\end{minipage} %
\begin{minipage}{0.51\linewidth}
    \centering
    \scriptsize
    \begin{tikzpicture}

    \newcommand{\myheight}{0.35\textwidth}

    \begin{axis}[
        legend pos=north west, %
        title style={font=\scriptsize},
        label style={font=\scriptsize},
        title style={yshift=-1.5ex},        
        title={},
        width=\textwidth,
        height=0.45\textwidth,
        ylabel={Density},
        xlabel={Cosine of angle},
        ytick=\empty,
        xmin=0, xmax=1,
        legend pos=north west,
        legend style={font=\scriptsize},        
    ]

        \addplot [name path=ZERO,draw=none, forget plot] coordinates {(0,0) (1,0)};
        
        \addplot[color=red, draw=none, name path=ID, forget plot]
        table[col sep=comma, x=x, y=y] {figure-data/projsims_id.csv};
        \addplot [blue, opacity=0.2] fill between [of=ID and ZERO];        
        
        \addplot[color=green, draw=none, name path=OOD, forget plot]
        table[col sep=comma, x=x, y=y] {figure-data/projmaxsims_id.csv};
        \addplot [cyan, opacity=0.2] fill between [of=OOD and ZERO]; 

        \legend{Angle to $W_\text{id}$, Angle within $W_\text{id}$};        
        
    \end{axis}

\end{tikzpicture} \vspace{-0.3cm}
    \captionof{figure}{Angles to $W_\text{id}$ vs. angles within $W_\text{id}$ to the closest class-mean for ID data.}%
    \label{fig:tosub-vs-insub}
\end{minipage}
\end{figure}

\subsection{Ablation: Alternative ID/OOD Decisions} \label{sec:decisions}

Several existing methods for OSSL use Otsu-thresholding to find a hard decision boundary for ID/OOD classification \cite{yu2020multi, huang2021trash, wang2023out}. To compare the Otsu approach \cite{otsu1979threshold} to our proposed probabilistic approach, we evaluate a version of ProSub where our probabilistic approach is replaced by a binary prediction given the Otsu threshold. Specifically, the Otsu threshold is evaluated at each training step given the subspace scores of the unlabeled data and is updated as an EMA. Additionally, we test a version of ProSub where our sampled binary mask for ID/OOD detection is replaced by the predicted probabilities directly. That is $m_i^\text{ood} - m_i^\text{id} \leftarrow 1 - 2p^\text{id}_i$ in \eqref{eq:betamodel-align} and the conditioning on $m_i^\text{id} = 1$ in \eqref{eq:betamodel-pl-loss} is replaced by scaling each term by $p^\text{id}_i$.

The results in \cref{tab:ablation} show that the Otsu approach performs significantly worse than the two probabilistic approaches. An examination of the Otsu run reveals that this approach assigns a too-low threshold early in training when the scores for ID and OOD data have a lot of overlap, resulting in many OOD data being predicted as ID. Note that the Otsu method inherently assumes $\pi=0.5$ \cite{kurita1992maximum}, corresponding to the true ID proportion in the evaluated scenario. The weighted version of ProSub, on the other hand, performs nearly identically to the version using sampled binary masks (as proposed in \cref{sec:l-sub}). This seems reasonable because, over a large number of training steps, the mean of the sampled masks should converge to the predicted probabilities.

\subsection{Ablation: Omitting Loss Terms}

To evaluate the influence of separate loss terms in ProSub, we conduct experiments where $\ell_\text{self}$ and or $\ell_\text{sub}$ are omitted. These experiments are run using CIFAR-100 as ID with 10,000 labels and CIFAR-10 as OOD. The results in \cref{tab:ablation} show that both $\ell_\text{self}$ and $\ell_\text{sub}$ separately contribute to higher AUROC and accuracy. Moreover, combining these loss terms yields the best overall performance.

\begin{table}[t]
    \scriptsize
    \centering
    \caption{Ablative experiments. \textbf{Left:} Evaluating alternative ID/OOD decisions in ProSub. ID: CIFAR-100 (2,500 labels), OOD: CIFAR-10. \textbf{Right:} Using subsets of loss terms in ProSub. ID: CIFAR-100 (10,000 labels), OOD: CIFAR-10.}
    \begin{tabular}{c c c}
\toprule
& Accuracy & AUROC \\ \midrule
ProSub (Otsu) & 70.75 & 0.67 \\
ProSub (weighted) & 74.01 & 0.97 \\
ProSub (unmodified) & 74.11 & 0.97 \\
\bottomrule
\end{tabular}
    \qquad
    \begin{tabular}{c c c c}
\toprule
$\ell_\text{self}$ & $\ell_\text{sub}$ & Accuracy & AUROC \\ \midrule
& & 70.24 & 0.63 \\
& \checkmark & 71.16 & 0.87 \\
\checkmark & & 78.26 & 0.96 \\
\checkmark & \checkmark & \textbf{79.46} & \underline{0.98} \\
\bottomrule
\end{tabular}
    \label{tab:ablation}
\end{table}

\section{Conclusion}
\label{sec:conclusion}

This work demonstrates that our proposed subspace score, based on computing angles between features and an ID subspace, is effective for ID/OOD classification in OSSL. Moreover, we show that the conditional distributions of scores given ID or OOD data can be estimated as Beta distributions through an iterative algorithm inspired by the Expectation-Maximization (EM) algorithm. The estimated conditionals enable probabilistic predictions of samples being ID or OOD. These components are used in the proposed ProSub, a method for OSSL that demonstrates state-of-the-art results on many benchmark datasets.

\section*{Acknowledgement}
\label{sec:acknowledgement}

This work was supported by Saab AB, the Swedish Foundation for Strategic Research, and Wallenberg AI, Autonomous Systems and Software Program (WASP) funded by the Knut and Alice Wallenberg Foundation. The computations were enabled by resources provided by the National Academic Infrastructure for Supercomputing in Sweden (NAISS), partially funded by the Swedish Research Council through grant agreement no. 2022-06725.

\bibliographystyle{splncs04}
\bibliography{main}

\clearpage
\setcounter{page}{1}
\begin{center}
    {\Large \textbf{ProSub: Probabilistic Open-Set Semi-Supervised Learning with Subspace-Based Out-of-Distribution Detection}} \\
    Supplementary Material
\end{center}

\section{Qualitative Analysis of Feature Separation}

To further analyze the effects of the self-supervision, $\ell_\text{self}$ from \eqref{eq:betamodel-self-supervision}, and the subspace loss, $\ell_\text{sub}$ from \eqref{eq:betamodel-align}, we plot t-SNE (Maaten and Hinton, 2008) reductions of features from ID and OOD test sets for a few different training setups. These experiments are done with CIFAR-100 as ID and CIFAR-10 as OOD. First, we train a fully supervised model using all 50,000 training data (with labels) from CIFAR-100. This model is never exposed to OOD data. Secondly, we train ProSub using 10,000 labels and first train until the end of the warm-up phase. At this stage, the model has only been trained with the labeled cross-entropy, $\ell_\text{sup}$ from \eqref{eq:betamodel-ll}, and self-supervision, $\ell_\text{self}$. Finally, we carry out a full training run of ProSub, where $\ell_\text{semi}$ from \eqref{eq:betamodel-pl-loss} and $\ell_\text{sub}$ are applied after the warm-up stage. %

\begin{figure}[]
    \centering
    \includegraphics[]{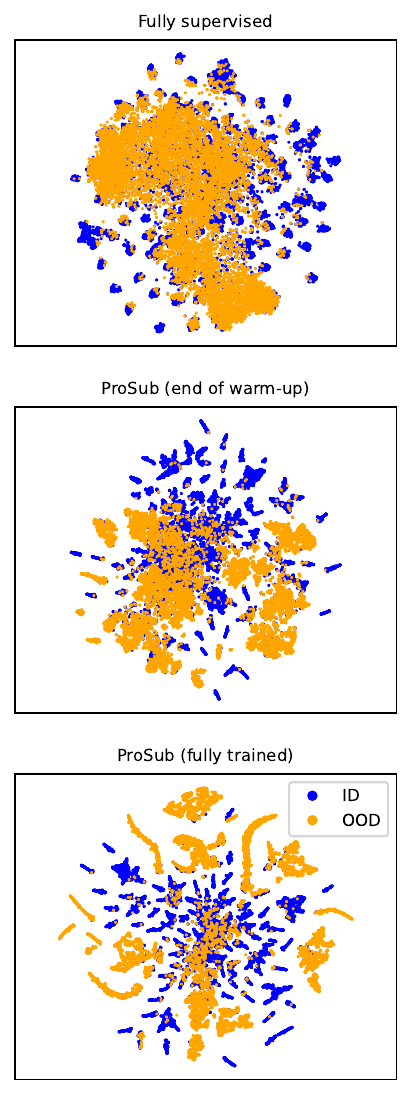}
    \caption{t-SNE of features. ID: CIFAR-100, OOD: CIFAR-10.}
    \label{fig:tsne}
\end{figure}

The results are shown in \cref{fig:tsne}. From the top panel, we see that the fully supervised model successfully clusters the ID data in feature space. However, most OOD data are not clustered or separated from ID, highlighting the challenge of OOD detection when we do not receive learning signals from these data. 

The next evaluated model is ProSub at the end of the warm-up. This model is trained with fewer labeled data than the fully supervised model but is exposed to both (unlabeled) ID and OOD through self-supervision. OOD data now begin to form distinct clusters, visibly separated from ID data. This suggests that self-supervision facilitates the clustering of both ID and OOD data. Visually, it seems reasonable to believe that OOD detection in this feature space is easier than for the fully supervised case. However, there are still regions where ID and OOD are mixed.

Finally, we have the features from the fully trained ProSub. Now we see even more clear and separated clusters for both ID and OOD data, indicating that the subspace loss further contributes to forming separated clusters for ID and OOD. An interesting observation is that OOD forms multiple clusters instead of one, even though this is not explicitly encouraged by either the self-supervision or the subspace loss. This indicates that the model not only learns to separate ID from OOD but also learns to group data within OOD.

\newpage

\section{Experiments with Unseen Outliers} \label{sec:unseen-ood}

\cref{tab:betamodel-results} evaluates AUROC of OOD detection on classes present in the unlabeled training set (seen OOD). While this is a core metric of OSSL performance, we also find value in exploring OOD detection for classes completely unseen during training (unseen OOD). To simulate this scenario, we divide Tiny ImageNet into three parts: 70 ID classes, 70 OOD classes present in the unlabeled training data, and 60 OOD classes entirely unseen during training. We use 3,500 labels. For this setting, we evaluate OpenMatch, SeFOSS, and ProSub. The results are shown in \cref{tab:unseen}. We see that ProSub drops in AUROC when going from seen to unseen OOD, indicating that the losses applied to OOD data facilitate learning features to discriminate between ID and seen OOD specifically. In contrast, OpenMatch obtains consistent AUROC for both seen and unseen OOD and SeFOSS shows \emph{better} AUROC for unseen OOD. However, despite this, ProSub demonstrates competitive results in OOD detection for unseen OOD.

\begin{table}[!t]
    \scriptsize
    \centering
    \caption{Evaluating OOD detection on unseen OOD using TIN.}    
    \begin{tabular}{c c c c}
\toprule
& & \multicolumn{2}{c}{AUROC} \\ \cmidrule{3-4}
& Accuracy & Seen OOD & Unseen OOD \\ \midrule
OpenMatch & 56.51 & 0.69 & 0.69 \\
SeFOSS &  64.09 & 0.68 & \underline{0.74} \\ 
ProSub & \textbf{66.06} & \underline{0.80} & 0.71 \\
\bottomrule
\end{tabular}
    \label{tab:unseen}
\end{table}

\section{Sensitivity Analysis of $\pi$} \label{sec:sensitivity-pi}

The probabilistic ID/OOD predictions of ProSub (see \eqref{eq:p-id}) require specifying the proportion of ID data in unlabeled data, $\pi$. In the experiments conducted for this work, we use exact values of $\pi$, which is $\pi=0.5$ for all scenarios except ImageNet20/10 where it is $\pi=0.66$. While it may be hard to know the exact value of $\pi$ in practice, we argue that it is easy to get an approximation by inspecting a subset of unlabeled data. If this approximation is unavailable, one can treat $\pi$ as a hyperparameter. To study how the performance of ProSub varies with $\pi$, we conduct experiments with CIFAR-100 as ID (10,000 labels) with CIFAR-10 as OOD using different values of $\pi$. With this setup, $\pi=0.5$ corresponds to the true proportion of ID data in unlabeled data.

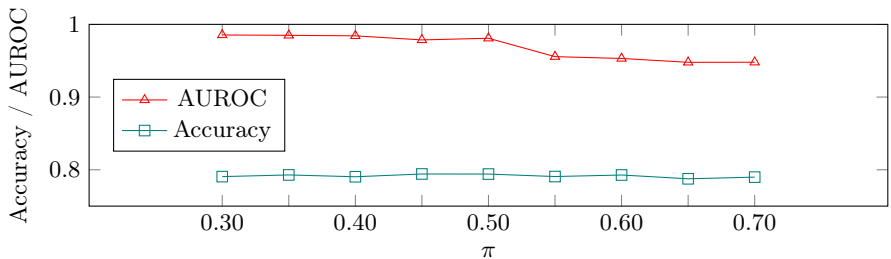
\begin{figure}[]
    \centering
    \begin{tikzpicture}
  \begin{axis}[
    title={},
    xlabel={$\pi$},
    ylabel={Accuracy / AUROC},
    legend style={at={(0.03,0.5)},anchor=west},   
    xtick={0.30, 0.35, 0.40, 0.45, 0.50, 0.55, 0.60, 0.65, 0.70},
    xticklabels={0.30, , 0.40, , 0.50, , 0.60, , 0.70},
    width=\columnwidth, %
    height=4cm, %
    xmin=0.20,
    xmax=0.80,
    ymin=0.75,
    ymax=1,
  ]

  \addplot[mark=triangle,red] coordinates {
    (0.3, 0.9854)
    (0.35, 0.9849)
    (0.40, 0.9842)
    (0.45, 0.9786)
    (0.50, 0.9809)
    (0.55, 0.9556)
    (0.60, 0.9531)
    (0.65, 0.9478)
    (0.70, 0.9479)
  };

  \addplot[mark=square,teal] coordinates {
    (0.3, 0.7906)
    (0.35, 0.7929)
    (0.40, 0.7904)
    (0.45, 0.7942)
    (0.50, 0.7941)
    (0.55, 0.7907)
    (0.60, 0.7928)
    (0.65, 0.7876)
    (0.70, 0.7899)
  };  
  
  \legend{AUROC, Accuracy}

  \end{axis}
\end{tikzpicture}
    \caption{Analyzing how ProSub performance depends on $\pi$ ($\pi=0.5$ corresponding to the true value).}
    \label{fig:pi-analysis}
\end{figure}

\cref{fig:pi-analysis} shows closed-set accuracy and AUROC as a function of $\pi$. We see that the obtained accuracy shows minimal dependency on $\pi$. The AUROC, interestingly, exhibits a stable high value as long as $\pi$ does not exceed 0.5. This suggests avoiding misclassifying OOD as ID is more crucial than the reverse. One possible explanation is that the cross-entropy for labeled data (or from pseudo-labeling) acts as an ``anchor'' for ID data, counter-acting the subspace loss that pushes these data away from $W_\text{id}$. No such counterweight exists if OOD data are pushed towards $W_\text{id}$, making this type of error more detrimental.

To show that we do not make significant performance gains from knowing the exact value of $\pi$, we include results on some datasets where ProSub displays the best results in \cref{tab:betamodel-results}. In these experiments, we use $\pi = 0.4$, \ie, lower than the true portion of ID data in the unlabeled data. These results are shown in \cref{tab:results-lowpi}, revealing that using an incorrect $\pi$ does not significantly impact our results. For TIN, the accuracy is slightly lower when using $\pi=0.4$, however, it is still higher than competing methods.

As a practical recommendation, we suggest using a $\pi$ slightly lower than the approximation obtained from unlabeled data to avoid exceeding the true proportion.

\begin{table}[]
    \scriptsize
    \centering
    \caption{Results from using an offset $\pi$: $\pi = 0.4$.}
    \begin{tabular}{c c c c}
\toprule
& ID: CIFAR-100 (10,000 lab.) & \multirow{2}{*}{IN50/50} & \multirow{2}{*}{TIN100/100} \\
& OOD: CIFAR-10 & & \\ \midrule
\multirow{2}{*}{ProSub (correct $\pi$)} & 79.59\spm{0.37} & 71.15\spm{0.80} & 60.92\spm{0.32} \\
& 0.98\spm{0.00} & 0.96\spm{0.00} & 0.72\spm{0.00} \\ \cdashlinelr{2-4}
\multirow{2}{*}{ProSub ($\pi = 0.4$)} & 79.54 & 71.48 & 59.96 \\
& 0.98 & 0.96 & 0.72 \\
\bottomrule
\end{tabular}
    \label{tab:results-lowpi}
\end{table}

\section{Hyperparameters}

The values of most hyperparameters used in ProSub are gathered from existing works and used without further tuning. For example, we use $w_\text{semi} = 1.0$ and initial learning rate $\eta_0 = 0.03$, $l^2$-regularization $w_\text{reg}$, decay rate $\gamma$, EMA momentum, batch sizes, and SGD momentum following \cite{sohn2020fixmatch, wallin2023improving}. For the evaluations done on TIN100/100 (new for this work), we copy the values for $w_\text{reg}$ and $\gamma$ used for CIFAR-100 in \cite{wallin2023improving} ($w_\text{reg} = 0.001$, $\gamma = 5/8$) because of the equal number of ID classes. For ImageNet50/50 (also new for this work) we copy the values for $w_\text{reg}$ and $\gamma$ used for ImageNet20/10 in \cite{wallin2023improving} ($w_\text{reg} = 0.0005$, $\gamma = 7/8$).

The main hyperparameter introduced for ProSub is $w_\text{sub}$, the weight for the subspace loss. We empirically find that $w_\text{sub} = 1.0$ works well across all evaluated datasets. Secondly, we use the cosine-based self-supervision from \cite{wallin2022doublematch} that shows $w_\text{self}$ can need dataset-specific tuning, which is why we use varying values of $w_\text{self}$.

\subsection{Selecting Hyperparameters Using Validation Data} \label{sec:hyperparams-validation}

The hyperparameters we tune for ProSub are $w_\text{self}$ and $w_\text{sub}$. Since labeled data are limited in OSSL, we suggest using a subset of labeled data as validation data to tune $w_\text{self}$ and $w_\text{sub}$. Subsequently, these tuned values can be utilized in a training run using all available labeled data for training.

We illustrate this procedure using CIFAR-100 as ID (10,000 labels) with CIFAR-10 as OOD by using 5,000 labels for training and 5,000 for validation. \Cref{tab:validation} shows that $w_\text{sub}=1.0$ and $w_\text{self} = 15.0$ yield the best validation accuracy among the evaluated values. Additionally, \cref{tab:validation} shows that these values correspond to the best accuracy on the test set. Notably, the closed-set accuracies align reasonably well with the obtained AUROC, simplifying hyperparameter selection as AUROC cannot be evaluated directly from the validation set.

The gap in accuracy between the validation set and the test set arises from labeled data (and consequently validation data) being included in the unlabeled training set without labels. To obtain an absolute prediction of test accuracy (rather than a relative one), the validation data can be explicitly excluded from the unlabeled set.

\begin{table}[]
    \scriptsize
    \centering
    \caption{Tuning hyperparameters from validation data.}
    \begin{tabular}{c c c c}
\toprule
\multicolumn{4}{c}{Validation results} \\ \midrule
& \multicolumn{3}{c}{$w_\text{self}$} \\ \cmidrule{2-4}
$w_\text{sub}$ & 5.0 & 15.0 & 25.0 \\ \cmidrule(lr){1-1} \cmidrule(lr){2-4}
0.1 & 86.82 & 87.58 & 88.12 \\
1.0 & 86.66 & \textbf{88.56} & 88.36 \\
10.0 & 52.00 & 79.44 & 85.84 \\ \bottomrule
\end{tabular}
\hspace{1cm}
\begin{tabular}{c c c c}
\toprule
\multicolumn{4}{c}{Test results} \\ \midrule
& \multicolumn{3}{c}{$w_\text{self}$} \\ \cmidrule{2-4}
$w_\text{sub}$ & 5.0 & 15.0 & 25.0 \\ \cmidrule(lr){1-1} \cmidrule(lr){2-4}
\multirow{2}{*}{0.1} & 72.67 & 75.72 & 75.08 \\
& 0.96 & 0.97 & 0.96 \\ \cdashlinelr{2-4}
\multirow{2}{*}{1.0} & 72.76 & \textbf{77.25} & 77.15 \\
& 0.86 & \underline{0.98} & \underline{0.98} \\ \cdashlinelr{2-4}
\multirow{2}{*}{10.0} & 12.25 & 58.78 & 71.43 \\
& 0.58 & 0.67 & 0.86 \\
\bottomrule
\end{tabular}%
    \label{tab:validation}
\end{table}

\subsection{The Number of Training Steps}

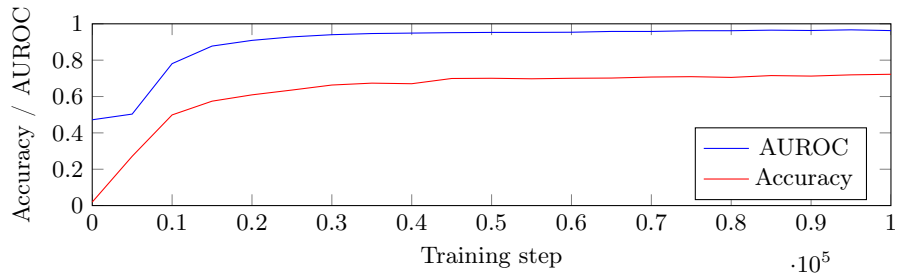
\begin{figure}[]
    \centering
    \begin{tikzpicture}
  \begin{axis}[
    title={},
    xlabel={Training step},
    ylabel={Accuracy / AUROC},
    legend pos=south east, %
    width=\linewidth, %
    height=4cm, %
    xmin=0,
    xmax=100000,
    ymin=0,
    ymax=1,
  ]

    \addplot[color=blue]
        table[col sep=comma, x=Step, y=Value]
        {figure-data/imagenet100-auroc.csv};

    \addplot[color=red]
        table[col sep=comma, x=Step, y=Value]
        {figure-data/imagenet100-acc.csv};

  \legend{AUROC, Accuracy}

  \end{axis}
\end{tikzpicture}
    \caption{ImageNet50/50 performance vs. training steps.}
    \label{fig:imagenet100-log}
\end{figure}

We set the number of training steps, $K$, to obtain reasonable training times, which is why we use a lower number of training steps for the ImageNet experiments. We have not observed any issues with overfitting or training collapse. The best performance is generally achieved at the end of training as shown in \cref{fig:imagenet100-log}. This figure shows test accuracy and AUROC as a function of training steps for a run on ImageNet50/50. This likely means that increasing the number of training steps should obtain equal or better results.

We have set the number of warm-up steps, $K_p$, to be a small but non-trivial fraction of the total number of training steps. \Cref{tab:kp} shows results on ImageNet50/50 with varying $K_p$ and a fixed $K=10^5$, showing that the results are insensitive to the choice of $K_p$.

\begin{table}[]
    \centering
    \caption{Varying $K_p$ on ImageNet50/50 (with $K=10^5$).}
    \label{tab:kp}
    \scriptsize
    \begin{tabular}{c c c c c c c} \toprule
       $K_p / 10^3$ & 15 & 20 & 25 & 30 & 35 & 40 \\ \midrule
        Acc & 71.92 & 72.52 & 71.44 & 71.15 & 71.40 & 71.60 \\
        AUROC & 0.96 & 0.96 & 0.96 & 0.96 & 0.96 & 0.96 \\ \bottomrule
    \end{tabular}
\end{table} 

\subsection{Fine-grained Hyperparameter Sensitivity} \label{sec:hyperparams-sensitivity}

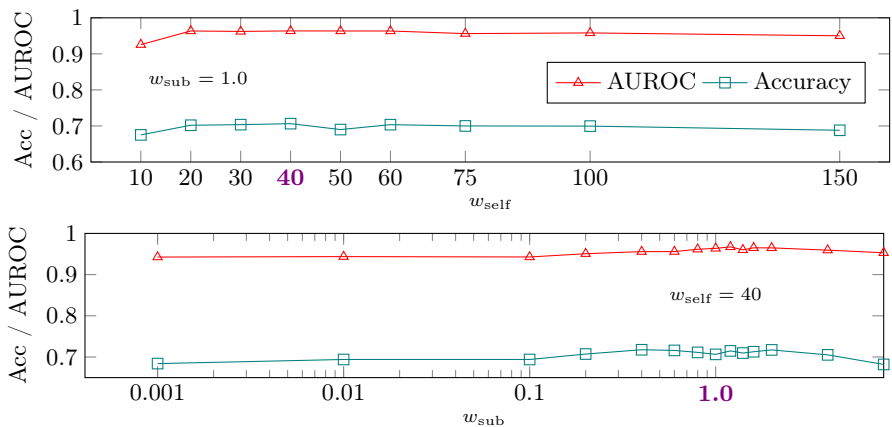
\begin{figure}[!h]
    \centering
    \scriptsize
    \begin{tikzpicture}
  \begin{axis}[
    title={},
    xlabel={$w_\text{self}$},
    ylabel={\footnotesize Acc / AUROC},
    x label style={at={(axis description cs:0.5,-0.2)},anchor=north},    
    ticklabel style = {font=\footnotesize},
    legend columns=-1,
    legend style={at={(0.97,0.55)},anchor=east},   
    xtick={10, 20, 30, 40, 50, 60, 75, 100, 150},
    xticklabels={10, 20, 30, \textcolor{violet}{\bf 40}, 50, 60, 75, 100, 150},
    width=\columnwidth, %
    height=3.5cm, %
    xmin=0,
    xmax=160,
    ymin=0.6,
    ymax=1,
  ]

  \addplot[mark=triangle,red] coordinates {
    (10, 0.9256)
    (20, 0.9635)
    (30, 0.9622)
    (40, 0.9637)
    (50, 0.9635)
    (60, 0.9634)
    (75, 0.9562)
    (100, 0.9581)
    (150, 0.9501)
  };

  \addplot[mark=square,teal] coordinates {
    (10, 0.6752)
    (20, 0.702)
    (30, 0.7036)
    (40, 0.7064)
    (50, 0.69)
    (60, 0.7036)
    (75, 0.70)
    (100, 0.6996)
    (150, 0.688)
  };  
  
  \legend{\footnotesize AUROC, \footnotesize Accuracy}

  \node[anchor=west] at (axis cs: 10,0.83) {$w_\text{sub}=1.0$};

  \end{axis}
\end{tikzpicture}

\begin{tikzpicture}
  \begin{axis}[
    title={},
    xlabel={$w_\text{sub}$},
    ylabel={\footnotesize Acc / AUROC},
    ticklabel style = {font=\footnotesize},
    x label style={at={(axis description cs:0.5,-0.2)},anchor=north},
    xtick={0.001, 0.01, 0.1, 1.0},
    xticklabels={0.001, 0.01, 0.1, \textcolor{violet}{\bf 1.0}},
    width=\columnwidth, %
    height=3.5cm, %
    xmin=0,
    xmax=8,
    xmode=log,
    ymin=0.65,
    ymax=1,
  ]

  \addplot[mark=triangle,red] coordinates {
    (0.001, 0.9426)
    (0.01, 0.9437)
    (0.1, 0.9428)
    (0.2, 0.9507)
    (0.4, 0.9558)
    (0.6, 0.9559)
    (0.8, 0.9615)
    (1.0, 0.9637)
    (1.2, 0.9673)
    (1.4, 0.9601)
    (1.6, 0.965)
    (2.0, 0.9649)
    (4.0, 0.9594)
    (8.0, 0.953)
  };

  \addplot[mark=square,teal] coordinates {
    (0.001, 0.684)
    (0.01, 0.694)
    (0.1, 0.694)
    (0.2, 0.7072)
    (0.4, 0.7176)
    (0.6, 0.716)
    (0.8, 0.7112)
    (1.0, 0.7064)
    (1.2, 0.7148)
    (1.4, 0.7096)
    (1.6, 0.7128)
    (2.0, 0.7172)
    (4.0, 0.7052)
    (8.0, 0.682)
  };

    \node[anchor=center] at (axis cs: 1,0.85) {$w_\text{self}=40$};

  \end{axis}

\end{tikzpicture}
    \caption{Hyperparameter evaluations on ImageNet50/50 test sets. \textcolor{violet}{\bf Violet} marks values used for \cref{tab:betamodel-results}.}
    \label{fig:hyperparams}
\end{figure}

To further analyze the sensitivity of hyperparameters $w_\text{sub}$ and $w_\text{self}$ we run experiments on ImageNet50/50 with varying $w_\text{sub}$ and $w_\text{self}$. \Cref{fig:hyperparams} shows that the results drop when we go far away from the values used to generate the main results in \cref{tab:betamodel-results}, but there are relatively large ranges for both $w_\text{sub}$ and $w_\text{self}$ where the results are stable.

\newpage

\subsection{Initiation of Beta Parameters}

For the IMM estimation, we use the initial guess $\alpha_\text{id} = \beta_\text{ood} = 10$, $\alpha_\text{ood} = \beta_\text{id} = 2$. We make this choice to ensure that the estimate for the ID distribution lies closer to 1.0 than the OOD distribution. However, because of the warm-up phase, the estimates have time to improve and settle before they are used to generate training signal through $\ell_\text{semi}$ \eqref{eq:betamodel-pl-loss} and $\ell_\text{sub}$ \eqref{eq:betamodel-align}. We have not found the initiation of these parameters to be significant for our performance.

\section{Varying ID/OOD Ratios in Unlabeled Data}

\begin{figure}[]
    \centering
    \begin{tikzpicture}

    \begin{groupplot}[
        group style={
            group size=1 by 2,
            x descriptions at=edge bottom,             
            horizontal sep=10pt,
        },
    ]
    \nextgroupplot[
        title style={yshift=-1.5ex},        
        title={ID: CIFAR-100 2,500 labels, OOD: CIFAR-10},
        width=\columnwidth,
        height=0.3\textwidth,
        ylabel={Acc / AUROC},
        legend columns=-1,
        legend style={
            at={(0.97,0.60)},
            anchor=east,},
        xmin=0, xmax=1,
    ]

      \addplot[mark=triangle,red] coordinates {
        (0.1, 0.80)
        (0.2, 0.87)
        (0.3, 0.89)
        (0.4, 0.96)
        (0.5, 0.96)
        (0.6, 0.97)
        (0.7, 0.97)
        (0.8, 0.96)
        (0.9, 0.95)
      };
    
      \addplot[mark=square,teal] coordinates {
      (0.1, 0.7143)
      (0.2, 0.7243)
      (0.3, 0.7407)
      (0.4, 0.7387)
      (0.5, 0.7338)
      (0.6, 0.7169)
      (0.7, 0.6939)
      (0.8, 0.6493)
      (0.9, 0.5995)
      };  
      
      \legend{AUROC, Acc}

    \nextgroupplot[
        title style={yshift=-1.5ex},
        title={ImageNet50/50},
        width=\columnwidth,
        height=0.3\textwidth,
        ylabel=Acc / AUROC,
        xlabel={Ratio of OOD in unlabeled data},
        xmin=0, xmax=1,
    ]

      \addplot[mark=triangle,red] coordinates {
        (0.1, 0.7827)
        (0.2, 0.8247)
        (0.3, 0.8982)
        (0.4, 0.9459)
        (0.5, 0.96)
        (0.6, 0.9343)
        (0.7, 0.9225)
        (0.8, 0.9207)
        (0.9, 0.922)
      };
    
      \addplot[mark=square,teal] coordinates {
      (0.1, 0.6868)
      (0.2, 0.6968)
      (0.3, 0.6972)
      (0.4, 0.7100)
      (0.5, 0.7204)
      (0.6, 0.6576)
      (0.7, 0.6056)
      (0.8, 0.5636)
      (0.9, 0.5476)
      };    

    \end{groupplot}

\end{tikzpicture}
    \caption{ProSub performance with varying ratios of ID and OOD in the unlabeled set.}
    \label{fig:ood-ratios}
\end{figure}
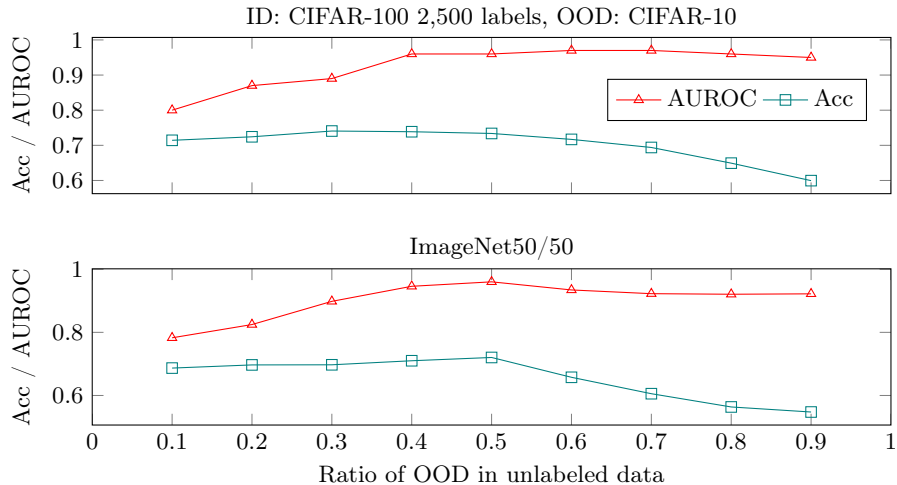

In the experiments of \cref{tab:betamodel-results}, most of our benchmark problems have equal amounts of ID and OOD in the unlabeled set. Here, we study how ProSub performs with varying ratios of ID to OOD data in the unlabeled set. \Cref{fig:ood-ratios} shows closed-set accuracy and AUROC for ProSub with varying OOD frequencies. We let $\pi$ follow the true ID/OOD ratio. For these experiments, we use CIFAR-100 (2,500 labels) as ID with CIFAR-10 as OOD, and ImageNet50/50. As expected, AUROC increases with more OOD data because the exposure to OOD data through self-supervision enables better OOD detection (see \cref{sec:enables-subspace}). Conversely, closed-set accuracy drops as ID data decreases due to fewer pseudo-labels that help us learn the ID classes. The results indicate optimal OOD frequencies around 0.4 - 0.5 that yield the best results for both OOD detection and closed-set accuracy. However, the OOD frequency is difficult to control in real-world scenarios.

\section{Regularization of ID Probabilities}

Based on the observation in \cref{sec:sensitivity-pi} that avoiding misclassifying ID as OOD is more important than the reverse, we find it beneficial to regularize the ID probabilities when computing the random mask in \eqref{eq:id-sampling}. This is achieved by adding a constant, $\epsilon$, to the denominator of \eqref{eq:p-id} as
\begin{equation} \label{eq:p-id-regularized}
    p(\mathbf{x} \in \mathcal{ID} | s(\mathbf{z})) = \frac{\pi p_\text{id}(s(\mathbf{z}))}{\pi p_\text{id}(s(\mathbf{z})) + (1 - \pi) p_\text{ood}(s(\mathbf{z})) + \epsilon}.
\end{equation}
We have found $\epsilon = 0.1$ to be a suitable value. Note that this regularization is used only for computing the random mask and not in the IMM estimation.

\section{Limitations}

\Cref{sec:unseen-ood} shows ProSub's superior performance on OOD detection for \emph{seen} OOD specifically. While ProSub remains competitive for unseen OOD detection, other methods may perform better if unseen OOD detection is your most important metric. Furthermore, this work only considers datasets that are balanced in terms of classes. We do not know how big shifts in class balances impact our performance. Finally, a limitation of ProSub lies in its dependence on dataset-specific tuning of $w_\text{self}$ and the necessity to tune $\pi$ or approximate the proportion of ID data within the unlabeled data.

\section{Score Distributions and Estimates}

\begin{figure}[]
    \centering
    \input{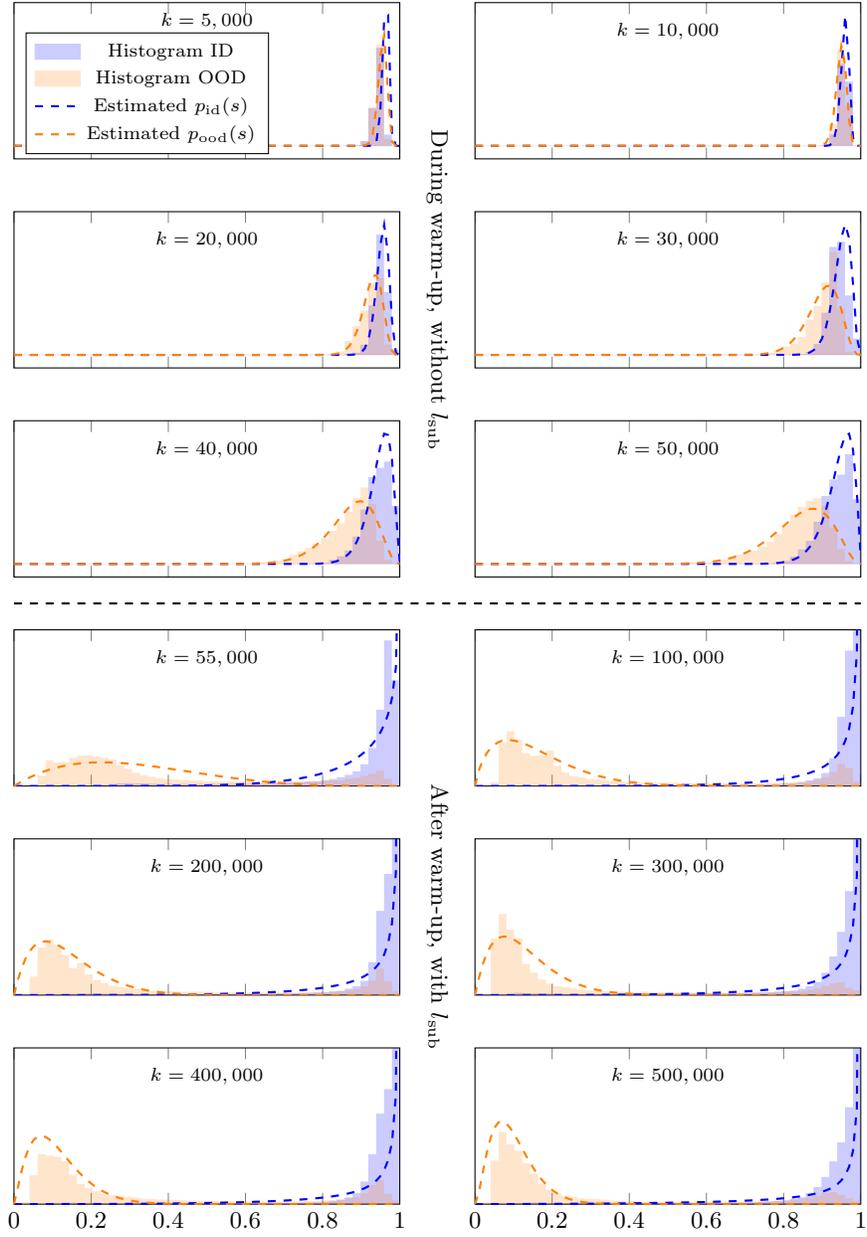}
    \caption{Distributions of scores and their corresponding estimates at different time steps during training.}
    \label{fig:histograms}
\end{figure}

In \cref{sec:results-densities} and \cref{fig:em-estimation} we look at the distributions of scores and the corresponding estimates at two different time steps during training. Here, in \cref{fig:histograms}, we show the equivalent evaluations at more time steps during training to display how the distributions and their corresponding estimates progress. These results are from a run using CIFAR-100 (2,500 labels) as ID with CIFAR-10 as OOD. The current training step is denoted by $k$ and the warm-up phase runs for 50,000 steps.

\Cref{fig:histograms} shows that during the warm-up phase, most data stay fairly close to $W_\text{id}$, but as training progresses, we start to distinguish between ID and OOD when the distribution of OOD moves slowly away from $W_\text{id}$. Interestingly, despite the overlapping distributions, the estimated Beta distributions accurately capture the individual mixture components throughout the warm-up phase.

After the warm-up phase (indicated by the horizontal black dashed line in \cref{fig:histograms}), when we apply $\ell_\text{sub}$ from \eqref{eq:betamodel-align}, we see that the distribution of scores for OOD data quickly moves away from $W_\text{id}$ (lower scores). The distribution of scores for ID data similarly moves closer to $W_\text{id}$ (higher scores). The estimated Beta distributions adapt well to this sudden change.

However, we also see that a few OOD data incorrectly get scores close to 1.0, highlighting that our obtained ID/OOD classifier does not have perfect accuracy. Notably, the set of OOD data that obtain high scores after the warm-up phase seems to grow and shrink in size at different time steps, indicating that the model can recover from misclassifying these data.

\newpage

\section{Indexing of Classes in TIN and IN100}

For completeness, we specify how we divide the classes of Tiny ImageNet and ImageNet100 into ID and OOD. How classes are indexed in ImageNet100 are shown in \cref{tab:in100-classes}. Here, we use indices 0-49 as ID and 50-99 classes as OOD.

The indexing of classes in Tiny ImageNet is shown in \cref{tab:tin-classes}. For experiments on TIN100/100, we use indices 0-99 as ID and 100-199 as OOD. For the experiments conducted using unseen OOD in \cref{sec:unseen-ood}, we use 0-69 ID, 70-139 as seen OOD, and 140-199 as unseen OOD. 

\newpage

\begin{table}[]
    \scriptsize
    \centering
    \caption{Class indexing for ImageNet100.}
    \begin{tabular}{c c}
\toprule
Class & Index \\ \midrule
n01440764 & 0 \\
n01443537 & 1 \\
n01484850 & 2 \\
n01491361 & 3 \\
n01494475 & 4 \\
n01496331 & 5 \\
n01498041 & 6 \\
n01514668 & 7 \\
n01514859 & 8 \\
n01531178 & 9 \\
n01537544 & 10 \\
n01560419 & 11 \\
n01582220 & 12 \\
n01592084 & 13 \\
n01601694 & 14 \\
n01608432 & 15 \\
n01614925 & 16 \\
n01622779 & 17 \\
n01630670 & 18 \\
n01632458 & 19 \\
n01632777 & 20 \\
n01644900 & 21 \\
n01664065 & 22 \\
n01665541 & 23 \\
n01667114 & 24 \\
n01667778 & 25 \\
n01675722 & 26 \\
n01677366 & 27 \\
n01685808 & 28 \\
n01687978 & 29 \\
n01693334 & 30 \\
n01695060 & 31 \\
n01698640 & 32 \\
n01728572 & 33 \\
n01729322 & 34 \\
n01729977 & 35 \\
n01734418 & 36 \\
n01735189 & 37 \\
n01739381 & 38 \\
n01740131 & 39 \\
n01742172 & 40 \\
n01749939 & 41 \\
n01751748 & 42 \\
n01753488 & 43 \\
n01755581 & 44 \\
n01756291 & 45 \\
n01770081 & 46 \\
n01770393 & 47 \\
n01773157 & 48 \\
n01773549 & 49 \\ \bottomrule
\end{tabular}
\begin{tabular}{c c}
\toprule
Class & Index \\ \midrule
n01773797 & 50 \\
n01774384 & 51 \\
n01774750 & 52 \\
n01775062 & 53 \\
n01776313 & 54 \\
n01795545 & 55 \\
n01796340 & 56 \\
n01798484 & 57 \\
n01806143 & 58 \\
n01818515 & 59 \\
n01819313 & 60 \\
n01820546 & 61 \\
n01824575 & 62 \\
n01828970 & 63 \\
n01829413 & 64 \\
n01833805 & 65 \\
n01843383 & 66 \\
n01847000 & 67 \\
n01855672 & 68 \\
n01860187 & 69 \\
n01877812 & 70 \\
n01883070 & 71 \\
n01910747 & 72 \\
n01914609 & 73 \\
n01924916 & 74 \\
n01930112 & 75 \\
n01943899 & 76 \\
n01944390 & 77 \\
n01950731 & 78 \\
n01955084 & 79 \\
n01968897 & 80 \\
n01978287 & 81 \\
n01978455 & 82 \\
n01984695 & 83 \\
n01985128 & 84 \\
n01986214 & 85 \\
n02002556 & 86 \\
n02006656 & 87 \\
n02007558 & 88 \\
n02011460 & 89 \\
n02012849 & 90 \\
n02013706 & 91 \\
n02018207 & 92 \\
n02018795 & 93 \\
n02027492 & 94 \\
n02028035 & 95 \\
n02037110 & 96 \\
n02051845 & 97 \\
n02058221 & 98 \\
n02077923 & 99 \\ \bottomrule
\end{tabular}
    \label{tab:in100-classes}
\end{table}

\newpage

\begin{table}[]
    \scriptsize
    \centering
    \caption{Class indexing for Tiny ImageNet.}    
    \begin{tabular}{c c}
\toprule
Class & Index \\ \midrule
n02814533 &	0 \\
n02113799 &	1 \\
n02883205 &	2 \\
n04597913 &	3 \\
n03733131 &	4 \\
n04179913 &	5 \\
n02802426 &	6 \\
n04070727 &	7 \\
n03706229 &	8 \\
n02321529 &	9 \\
n02085620 &	10 \\
n03970156 &	11 \\
n02730930 &	12 \\
n02268443 &	13 \\
n02099712 &	14 \\
n04133789 &	15 \\
n04251144 &	16 \\
n03026506 &	17 \\
n04532106 &	18 \\
n07614500 &	19 \\
n07747607 &	20 \\
n01742172 &	21 \\
n03160309 &	22 \\
n03992509 &	23 \\
n01784675 &	24 \\
n01644900 &	25 \\
n02808440 &	26 \\
n01774750 &	27 \\
n02669723 &	28 \\
n03838899 &	29 \\
n01910747 &	30 \\
n03444034 &	31 \\
n04118538 &	32 \\
n03662601 &	33 \\
n02948072 &	34 \\
n02231487 &	35 \\
n02106662 &	36 \\
n02094433 &	37 \\
n07873807 &	38 \\
n01641577 &	39 \\
n03977966 &	40 \\
n04259630 &	41 \\
n07871810 &	42 \\
n02906734 &	43 \\
n02364673 &	44 \\
n04008634 &	45 \\
n09256479 &	46 \\
n02815834 &	47 \\
n02481823 &	48 \\
n02963159 &	49 \\ \bottomrule
\end{tabular}
\begin{tabular}{c c}
\toprule
Class & Index \\ \midrule
n03100240 &	50 \\
n04149813 &	51 \\
n01917289 &	52 \\
n04507155 &	53 \\
n02892201 &	54 \\
n03089624 &	55 \\
n02132136 &	56 \\
n04254777 &	57 \\
n02927161 &	58 \\
n03983396 &	59 \\
n02123045 &	60 \\
n02791270 &	61 \\
n09246464 &	62 \\
n03447447 &	63 \\
n04417672 &	64 \\
n07579787 &	65 \\
n07583066 &	66 \\
n02795169 &	67 \\
n03393912 &	68 \\
n04023962 &	69 \\
n04486054 &	70 \\
n02233338 &	71 \\
n01855672 &	72 \\
n02814860 &	73 \\
n04067472 &	74 \\
n02410509 &	75 \\
n02480495 &	76 \\
n03126707 &	77 \\
n07753592 &	78 \\
n03085013 &	79 \\
n02988304 &	80 \\
n02099601 &	81 \\
n04501370 &	82 \\
n02909870 &	83 \\
n03014705 &	84 \\
n04146614 &	85 \\
n02666196 &	86 \\
n04074963 &	87 \\
n01882714 &	88 \\
n03930313 &	89 \\
n07734744 &	90 \\
n04366367 &	91 \\
n03837869 &	92 \\
n03250847 &	93 \\
n02236044 &	94 \\
n03201208 &	95 \\
n02437312 &	96 \\
n02837789 &	97 \\
n02699494 &	98 \\
n04099969 &	99 \\ \bottomrule
\end{tabular}
\begin{tabular}{c c}
\toprule
Class & Index \\ \midrule
n07615774 &	100 \\
n03355925 &	101 \\
n04371430 &	102 \\
n01945685 &	103 \\
n03649909 &	104 \\
n03404251 &	105 \\
n03891332 &	106 \\
n07695742 &	107 \\
n04311004 &	108 \\
n02823428 &	109 \\
n07749582 &	110 \\
n04399382 &	111 \\
n07875152 &	112 \\
n09193705 &	113 \\
n02074367 &	114 \\
n03937543 &	115 \\
n02206856 &	116 \\
n01698640 &	117 \\
n02788148 &	118 \\
n02917067 &	119 \\
n01983481 &	120 \\
n02504458 &	121 \\
n02281406 &	122 \\
n04376876 &	123 \\
n02056570 &	124 \\
n03388043 &	125 \\
n02423022 &	126 \\
n07720875 &	127 \\
n02125311 &	128 \\
n03400231 &	129 \\
n02226429 &	130 \\
n04465501 &	131 \\
n02841315 &	132 \\
n02843684 &	133 \\
n09332890 &	134 \\
n02415577 &	135 \\
n04596742 &	136 \\
n04275548 &	137 \\
n01774384 &	138 \\
n02793495 &	139 \\
n02395406 &	140 \\
n07715103 &	141 \\
n03255030 &	142 \\
n02403003 &	143 \\
n04456115 &	144 \\
n04398044 &	145 \\
n12267677 &	146 \\
n03424325 &	147 \\
n01950731 &	148 \\
n01984695 &	149 \\ \bottomrule
\end{tabular}
\begin{tabular}{c c}
\toprule
Class & Index \\ \midrule
n01768244 &	150 \\
n03617480 &	151 \\
n04487081 &	152 \\
n07768694 &	153 \\
n02002724 &	154 \\
n06596364 &	155 \\
n03042490 &	156 \\
n04285008 &	157 \\
n03544143 &	158 \\
n03980874 &	159 \\
n02279972 &	160 \\
n03770439 &	161 \\
n04560804 &	162 \\
n07711569 &	163 \\
n04356056 &	164 \\
n02977058 &	165 \\
n03854065 &	166 \\
n03179701 &	167 \\
n02486410 &	168 \\
n02058221 &	169 \\
n09428293 &	170 \\
n04265275 &	171 \\
n01443537 &	172 \\
n03814639 &	173 \\
n02165456 &	174 \\
n02129165 &	175 \\
n02509815 &	176 \\
n02190166 &	177 \\
n02124075 &	178 \\
n07920052 &	179 \\
n03804744 &	180 \\
n01770393 &	181 \\
n04562935 &	182 \\
n03976657 &	183 \\
n04328186 &	184 \\
n03599486 &	185 \\
n02999410 &	186 \\
n03637318 &	187 \\
n03584254 &	188 \\
n02769748 &	189 \\
n02123394 &	190 \\
n04540053 &	191 \\
n03763968 &	192 \\
n03902125 &	193 \\
n03670208 &	194 \\
n03796401 &	195 \\
n01629819 &	196 \\
n02950826 &	197 \\
n04532670 &	198 \\
n01944390 &	199 \\ \bottomrule
\end{tabular}%
    \label{tab:tin-classes}
\end{table}

\end{document}